\documentclass[A4]{IEEEtran}
\IEEEoverridecommandlockouts
\usepackage{cite}

\usepackage{amsmath,amssymb,amsfonts}
\usepackage{amssymb}
\usepackage{breqn}
\usepackage{cleveref}
\usepackage[lined,ruled,linesnumbered,commentsnumbered]{algorithm2e}
\usepackage{booktabs}
\usepackage{cases}
\usepackage{graphicx}
\usepackage{multirow}
\usepackage{mathrsfs}
\usepackage{textcomp}
\usepackage{xcolor}
\usepackage{subfigure}
\usepackage{boxedminipage}
\usepackage{threeparttable}
\usepackage{stfloats}
\usepackage{inputenc}

\def\BibTeX{{\rm B\kern-.05em{\sc i\kern-.025em b}\kern-.08em
    T\kern-.1667em\lower.7ex\hbox{E}\kern-.125emX}}

\begin{document}


\title{Evolutionary Gait Transfer of Multi-Legged Robots in Complex Terrains}
\author{Min JIANG,~\IEEEmembership{Senior~Member,~IEEE,}
	Guokun~CHI,~\IEEEmembership{Student~Member,~IEEE,}
	Geqiang PAN,~\IEEEmembership{}
	Shihui~GUO,~\IEEEmembership{Member,~IEEE,}	
	and Kay Chen TAN,~\IEEEmembership{Fellow,~IEEE,}
	\IEEEcompsocitemizethanks{\IEEEcompsocthanksitem M. JIANG, G. CHI, S. GUO and G. PAN are with the school of informatics, Xiamen University, China, Fujian, 361005.
		\IEEEcompsocthanksitem K. C. TAN is with the Department of Computer Science, City University of Hong Kong.}
	
}

\bibliographystyle{IEEEtran}

\maketitle

\begin{abstract}
Robot gait optimization is the task of generating an optimal control trajectory under various internal and external constraints. Given the high dimensions of control space, this problem is particularly challenging for multi-legged robots walking in complex and unknown environments. Existing literatures often regard the gait generation as an optimization problem and solve the gait optimization from scratch for robots walking in a specific environment. However, such approaches do not consider the use of pre-acquired knowledge which can be useful in improving the quality and speed of motion generation in complex environments. To address the issue, this paper proposes a transfer learning-based evolutionary framework for multi-objective gait optimization, named Tr-GO. The idea is to initialize a high-quality population by using the technique of transfer learning, so any kind of population-based optimization algorithms can be seamlessly integrated into this framework. The advantage is that the generated gait can not only dynamically adapt to different environments and tasks, but also simultaneously satisfy multiple design specifications (e.g., speed, stability). 
The experimental results show the effectiveness of the proposed framework for the gait optimization problem based on three multi-objective evolutionary algorithms: NSGA-II, RM-MEDA and MOPSO.
When transferring the pre-acquired knowledge from the plain terrain to various inclined and rugged ones, the proposed Tr-GO framework accelerates the evolution process by a minimum of 3-4 times compared with non-transferred scenarios.
\end{abstract}


\section{INTRODUCTION}

Multi-legged robots have attracted a wide range of researchers' interests due to their superior capability to maintain balance and walk in rugged terrains \cite{todd2013walking}. However, this ability can only be accomplished with the appropriate motions, so the quick and accurate generation of gaits for a multi-legged robot has become one of the most important prerequisites to complete its tasks.
Different approaches have been proposed to solve the problem of gait generation. For example, in \cite{jeong2019robust}, the authors proposed a motion controller for a biped robot that optimized different push recovery strategies, and the method can generate appropriate gaits by determining how the robot should respond to an external push. In \cite{juang2017multiobjective}, the authors proposed a fully connected recurrent neural network (FCRNN) for solving the multi-objective gait generation. The FCRNN can be considered as a central pattern generator and is optimized to produce angles of different degrees of freedom (DOFs). In \cite{boussema2019online}, the authors quantified leg capabilities and then proposed leg utility for coordinating adaptive lift-off and touch-down of stance legs. 

However, generating high-quality poses for multi-legged robots still remains a challenging task for two reasons. First, due to the large number of DOFs of multi-legged robots, it is difficult to quickly determine a feasible trajectory from the huge search space. Second, the trajectory to be executed should satisfy multiple requirements which are potentially contradictory. For example, the robot should not only be able to navigate through dangerous areas as quickly as possible (speed), but also remain upright and avoid falling over (stability); in addition, we want the robot to function for as long as possible given a limited amount of power consumption (energy).

Among the existing methods, the gait generation algorithms based on population optimization techniques have unique advantages  \cite{gong2010review}. The reason is that gait generation is often transformed into the task of solving multi-objective, multi-constraint optimization problems.
The traditional analytical methods are rarely effective when applied to such problems, while the population-based optimization method can produce multiple candidate motions simultaneously, allowing the controller to select the appropriate solution and complete specific tasks. For example, in \cite{chernova2004evolutionary}, the authors proposed an evolutionary approach to optimize forward gaits of quadruped robots; this method is more robust than the previous algorithms and also significantly increases the speed of problem solving. In \cite{ calandra2016bayesian, rai2018bayesian}, the authors proposed a Bayesian optimization algorithm for controller parameter optimization in order to reduce the need for human expertise and time-consuming design processes. In \cite{ degrave2015transfer }, the authors proposed to use a particle swarm algorithm for gait learning. To decrease learning time, the authors used the best 20 particles across all generations from a previous optimization to initialize the particle population. Experiments have shown that this simple strategy can be effective in helping robots learn new motion skills.

Multi-objective evolutionary algorithms (MOEAs) are often computationally-intensive and thus time-consuming.
An ill-initialized population could lead to a prolonged evolution process. 
However, most of the existing population-based optimization methods \cite{gong2010review, wright2015intelligent, he2019survey} do not have an effective mechanism to generate a high-quality initial population. 
It should be noted that the previous method \cite{ degrave2015transfer } directly copied the individuals with the best past performances to initialize a new population, based on the assumption that the pre-obtained actions and the actions to be learned share the identical distribution. Unfortunately, in many real situations, the actions that have been learned are not the same as the actions that will be learned. For example, walking on the ground and walking on a slope full of obstacles are not exactly the same, although there is a certain level of implicit similarity between the two. If we directly use the individuals with the best past performances to obtain the initial population, the optimization will likely be led in the wrong direction, and thus, the solution speed will be slower.
Similar problems, namely, training and testing samples do not follow the independent and identical distribution\cite{8100935}, are also of considerable concern in the field of machine learning. Meanwhile, in order to solve these problems, different transfer learning methods \cite{weiss2016survey,9097186} have been proposed and proven to be powerful tools.

In this paper, we consider the problem of robot gait generation as a multi-objective optimization problem (MOP), and on this basis, we propose a method to solve this problem by combining transfer learning with the population-based optimization algorithm. Specifically, we propose an algorithm framework to obtain an initial population in a new environment by re-using the obtained optimal gaits from an old one.
Through this initial population, any evolutionary multi-objective optimization algorithm can be used to obtain new motions to meet multiple design specifications (e.g., speed, stability).
The advantage of the proposed method is to generate optimal gait patterns that can satisfy multiple requirements simultaneously.

The contribution of this research is that we consider the gait generation problem as an MOP, and propose a solution framework combining transfer learning and MOEAs to solve this problem. 
This framework not only enables the obtained motion to satisfy the requirements of multiple objective functions simultaneously, but also significantly improves the solving speed by reusing latent knowledge obtained in the past to solve problems in the current environment.
Our Tr-GO framework accelerates the evolution process by a minimum of 3-4 times compared with non-transferred scenarios.

The rest of this paper is organized as follows: Sec. II will present the preliminary information and existing relevant work. In Sec. III, we will introduce the robot model, two optimization goals and transfer learning modular. Subsequently, we will propose an algorithm framework for robot gait optimization based on transfer learning. Sec. IV presents the experimental results of our framework combined with three MOEAs in three different terrain environments. Sec. V summarizes our work and proposes directions for  future research.

\section{PRELIMINARY AND RELATED WORKS}
\subsection{Multi-Objective Optimization}

A general multi-objective optimization problem (MOP) is defined as follows:
\begin{align}\label{eq:multiobj}
& \textrm {Minimize } \mathbf {F(x)} =[F_1(x),F_2(x),...,F_k(x)]^T  \nonumber\\
& \textrm {subject to } g_j(x) \leq 0, j = 1,2,...,m, \\
& h_l(x)=0, l = 1,2,...,e,  \nonumber
\end{align}

where $k$ is the number of objective functions, $m$ is the number of inequality constraints $g(x)$, and $e$ is the number of equality constraints $h(x)$. $x \in E^n$ is a vector of decision variables, where $n$ is the number of independent variables $x_i$. $F (x) \in E^k $ is a vector of objective functions $F_i(x): E^n \rightarrow E^1$. $F_i(x)$ are also called objectives, criteria, payoff functions, cost functions, or value functions \cite{marler2004survey}.

\textbf{Definition 1}. [Decision Vector Domination] A decision vector $x_1$ Pareto dominates another vector $x_2$, denoted by $x_1\succ x_2$, if and only if:
\begin{equation}\label{eq:domination}
\left\{
\begin{array}
{lr}
\forall i = 1,...,k,F_i(x_1) \leq F_i(x_2) \\
\exists i = 1,...,k,F_i(x_1) < F_i(x_2) \\
\end{array}
\right.
\end{equation}

\textbf{Definition 2}. [Pareto Optimal Set] Both $x$ and $x^*$ are decision vectors, and if a decision vector $x^*$ is said to be non-dominated if and only if there is no other decision vector $x$ such that $x \succ x^* $. The Pareto optimal set (POS) \cite{pareto1906manuale,Athan1996A} is the set of all Pareto optimal solutions, that is:
\begin{equation}\label{eq:pareto}
POS = \{ x^* | \nexists x, x \succ x^*\}
\end{equation}

\textbf{Definition 3}. [Dynamic Pareto-optimal Front] The function space representation of all the solutions in the Pareto optimal set is the Pareto optimal front (POF).
\begin{equation}\label{eq:dynamic}
POF = \{ F(x^*) | x^* \in POS \}
\end{equation}

When there are two, three or more objectives, the POF is represented as a curve, surface or hyper-surface respectively.
Take a two-objective optimization as an example.
Two different decision variables are related to each other in two possible ways: either one dominates the other, or none of them is dominated \cite{deb2001multi}.
This creates a partially-ordered search space and a challenging optimization landscape.
%

\subsection{Related Work}
Gait generation is important in the field of robotics research.
Even with viable controller parameterization, finding the optimal parameter values is difficult. 
This parameter optimization often requires specific expert knowledge and a large number of robot experiments.  
Most of the existing methods follow two strategies to obtain the natural and effective gait of legged robots \cite{gong2010review}. The first strategy is bio-inspired in that it takes advantage of the biological heuristics of human and animal to complete the tasks of gait optimization for robots. This view follows the Darwinian theory of evolution: human and animal's viability consistently enforces their activities to follow the principle of minimizing metabolic costs \cite{mcneill2002energetics, bertram2005constrained,Manoj2006Computer}. However, the reality is that due to the inconsistency in kinematics and dynamics between humans/animals and robots, there is still a large gap to directly apply bio-motion data to robots. 
The second strategy is based on a heuristic optimization method, since the search for the desired gait parameters can be defined as an optimization problem. When combined with appropriate optimization methods, the problem is converted into an automatic search process for optimal gait parameters. At present, different methods \cite{hereid2018dynamic, gong2010review,silva2016open,takemori2018gait} have been proposed for automatic gait optimization. 
As a branch of the evolutionary robot (ER) field \cite{nolfi2016evolutionary}, gait evolution with the heuristic evolutionary algorithm has attracted much attention. 

The field of ER studies the automatic synthesis of robot control and/or hardware based on the principle of natural evolution. This process is similar to its real equivalence in that it uses evolutionary computation to generate robots that adapt to the environment \cite{floreano2008evolutionary}. 
Many researchers have applied evolutionary computation (EC) to the field of gait generation and optimization \cite{doncieux2013behavioral}. 
Previous methods used EC to optimize the four-legged Sony AIBO robot \cite{chernova2004evolutionary,tellez2006evolving}, or to solve the problem of recovery and optimization of the gait after the loss of the legged robot \cite{bongard2008accelerating}. The most commonly-used EC method for solving gait optimization is the genetic algorithm (GA). It can be applied to optimization problems for specific gaits \cite{Cardenasmaciel2011Generation,eaton2015evolutionary}. Especially in the research field of biped robots, GA is often used for solving optimization problems with dynamic objective functions \cite{gupta2018trajectory}. 
Researchers used GA to generate an energy-efficient trajectory for the compliant link biped, enabling it to complete tasks including walking with different step lengths and on variable slopes \cite{sarkar20158}.
Other successful applications of GAs included generating energy-efficient walking motion for 3-DOF robots  \cite{cardenas2011generation} and stair-climbing movements for a 12-DOF biped robot \cite{lim2014gait}. Other EC methods are also widely used in the motion optimization process of robots. 
Among them, the particle swarm optimization (PSO) algorithm was applied to identify the optimal control pattern for fast swimming \cite{wang2017autonomous}, and realized the speed of movement beyond their previous work. The distributed estimation algorithm (EDA) was used to solve the motion generation optimization problem of multi-legged robots and other tasks in complex environments \cite{jiang2017motion,9262190}.
In this paper, we propose a framework that is based on multi-objective optimization and can be combined with any population-based algorithms mentioned above to solve the gait generation problem. 

The gait optimization process requires a reasonable trade-off between performance metrics (e.g., speed of motion, stability criteria, actuation force, and energy consumption) in order to complete the desired task. Therefore, the gait generation problem can be formulated as an MOP by defining multiple specifically-designed objective functions \cite{gong2010review}. Different from the single-objective optimization problem, MOP is more suitable for more complex  tasks of robot gait generation. In \cite{nygaard2017multi}, researchers used a high-level controller that concurrently optimizes two objectives of speed and stability to achieve a series of robust gaits for quadruped robots. Another work \cite{kobayashi2015selection} proposed a motion selection algorithm that is capable of autonomously selecting the best motion to improve overall mobility. Based on this method, two strategies were proposed, namely gait selection and adaptive movement speed. These enhanced the robot's ability to maintain a fast moving speed in an unstable environment. Through the trade-off between the two goals, the performance of MOP on physical robots is better than that of single-objective optimization. 
\cite{raj2019multiobjective} proposed a new analysis method to optimize the trade-off stability and energy function of the biped robot. The author used a Real Coded Genetic Algorithm (RCGA) \cite{Nasu2002Optimal} to optimize the walking function of the robot to produce a set of optimal walking parameters. 
MOEA was used to optimize parameters of the walking pattern and improved the energy efficiency and stability of NAO robots by 67.05\% and 75\%, respectively. 
Another work \cite{moore2016comparison} compared the performance of two MOEAs: Lexicase selection and NSGA-II\cite{Kalyanmoy2002fast} in solving the task of gait evolution of quadruped robots. 
The objective functions include traveling distance, efficiency, and vertical torso movement.
For comparison, a baseline is optimized solely on a single objective: traveling distance. 
The results showed that the NSGA-II algorithm significantly outperforms Lexicase selection in all three functions, while Lexicase selection significantly outperforms the baseline in two of the three functions.

Traditional evolution techniques typically optimized for specific motion modes (such as forward motion). Recently, studies have shown that quality diversity \cite{pugh2016quality} can generate a variety of motion behaviors for complex robots in each evolutionary operation \cite{cully2013behavioral, cully2016evolving, cully2015robots}. The goal of a mass diversity algorithm is to create a solution archive that shows different behaviors or characteristics. Researchers also proposed Behavior Repertoire (BR) Evolution \cite{cully2013behavioral} to learn a large number of actions without the need to test each action individually. However, this algorithm lacks the possibility of autonomous exploration and proxy capabilities. 
Furthermore, a transferability-based BR-Evolution \cite{cully2016evolving} algorithm proposed a new evolutionary scheme that can simultaneously discover simple walking controllers that contain multiple directions. The algorithm uses a novel strategy with local competition to search for high-performance and diversified solutions, while retaining potential solutions that are usually discarded during evolution, so that the algorithm can quickly obtain more actions in the reachable space. Despite their success in tasks such as moving the robot forward, these quality diversity evolutionary algorithms still face great challenges when solving more complex tasks or adapting to more complex terrains. From motion control to task-oriented control, it is necessary to coordinate all the motion parameters, and adapt to the robot's sensory input and task goals. 
To enable ER technology to be used in complex robots beyond simple motion tasks, an evolutionary library-based control (EvoRBC) method  was proposed \cite{Duarte2018Evolution}. EvoRBC uses the diversity-driven algorithm to develop a complete function table of low-level primitives. At the same time, the neural evolution algorithm NEAT is used to evolve a high-level controller. The EvoRBC algorithm allows the robot to flexibly cope with complex tasks in real-world scenarios. Since the evolution performance of the algorithm is directly related to the design dimensionality of the evolutionary neural network, it is inevitable to increase time and calculation cost.

Although the application of evolutionary algorithms to robots has been successful to some extent, using these methods when dealing with new and complex tasks often require excessive time and computational costs in order to perform the necessary evolutionary steps. 
In contrast, human learners quickly adapt to new problems by transferring knowledge from relevant tasks in previous experiences.
This is known as \emph{transfer learning}.
In the domain of evolving gait patterns for multi-legged robots, little consideration is given to improving optimization efficiency by reusing the pre-acquired knowledge and reducing unnecessary computational overhead during the experiment. 
Therefore, it is the goal of this work to introduce transfer learning to accelerate optimization in a new environment\cite{9122031,9185798,9199822,9002942}.

\section{THE PROPOSED ALGORITHM FRAMEWORK}
\subsection{Multi-Legged Robot Model}
\begin{figure}[h]
	\centering
	\includegraphics[width=\linewidth]{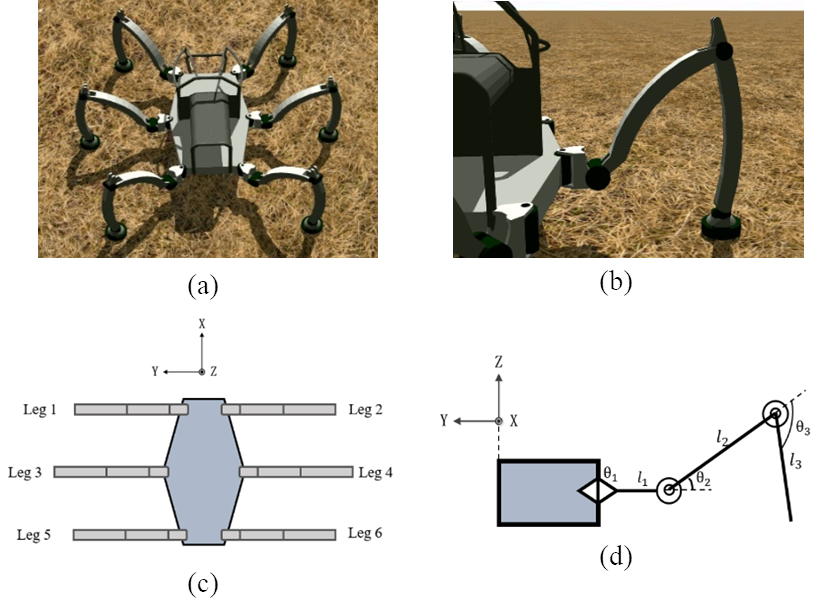}
	\caption{Illustration of the mantis hexapod robot model. (a) top view, (b) close view of a single leg, (c) schematic diagram of the hexapod robot model, (d) the rotation angles of a single leg.}
	\label{fig:robot}
\end{figure}
\begin{figure}[h]
	\centering
	\includegraphics[width=\linewidth]{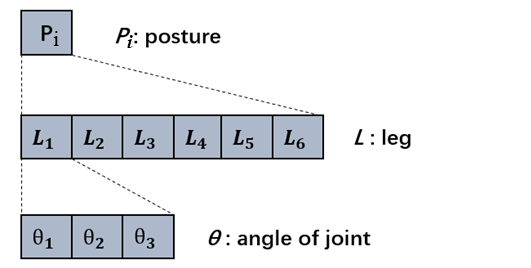}
	\caption{Representation of control trajectories for an individual robot.}
	\label{fig:trajectoy}
\end{figure}

In this paper, we use the mantis hexapod robot model (Fig.~\ref{fig:robot}) in the simulation software Webots \cite{michel2004cyberbotics}. To facilitate the description, its structure can be simplified as shown in Fig.~\ref{fig:robot}(c). The legs, identified as No. 1 to 6, are fixed outside the body at equal intervals. 
The robot posture is determined by the leg joints. 
The robot has 18 DOFs, i.e., three joints per leg. Each joint is a hinge that allows movement along the pivot axis. 
In Fig.~\ref{fig:robot}(d), $\theta_1$ represents the rotation angle around the $z$ axis, and $\theta_2$ and $\theta_3$ are the rotation angles around the $x$ axis. 
The foot position $P = \left( p_x, p_y, p_z \right )$ is determined by the rotation angle of each joint and the length of each link, and can be calculated by the forward kinematics equation \cite{iwasa2016motion}:
\begin{align}\label{eq:fk}
P = Trans(0,l_0,0) \cdot &  Rot(z, \theta _{m,1}) \cdot Trans(0,l_1,0) \cdot \nonumber \\
& Rot(x, \theta _{m,2}) \cdot Trans(0,l_2,0) \cdot  \\
& Rot(x, \theta _{m,3}) \cdot Trans(0,l_3,0) \cdot \mathbf{u} \nonumber
\end{align}
where $Trans\left (·,·,·\right )$ and $Rot \left( ·,· \right )$ denote the translation and rotation matrix respectively. $l_i$ is the length of the $i^{th}$ link and $\theta_{m,i}$ is the rotation angle of the $i^{th}$ joint in the $m^{th}$ leg, $\mathbf{u}$ stands for the normal vector. In this way, the walking behavior can be seen as performing a sequence of robot postures, each of which is a vector of joint angles (Fig.~\ref{fig:trajectoy}).  

Each joint on the leg is actuated with a proportional–integral–derivative (PID) actuator.
The target position of the PID actuator follows a template of sinusoidal function: $\theta_j = a_j\ \text{sin}(\omega t + p_j) + d_j$.
$a_j, p_j, d_j$ are the amplitude, phase offset and drifting values for the $k^{th}$ joint.
These variables are to be optimized in the evolutionary algorithm.
During the evolutionary process, the values of joint angles are within the operational range of robot joint on each leg.

Considering the popularity used in multi-legged robots, the triangular gait pattern is selected as a template gait. 
To perform the triangular gait \cite{guo2014locomotion}, the front and back legs on one side and the middle leg on the opposite side constitute as one group and offer a stable triangular support. 
Two groups of legs alternately lift up and sequentially move forward, which enables insects to move rapidly.
From the perspective of bionics, insects often move in an alternating triangular gait across a diverse range of speeds and terrains. 
This gait pattern is selected as the initial value for generating random gait samples.

\subsection{Optimization Objectives}
We consider the gait optimization for a multi-legged robot as a multi-objective problem. Two objective functions, walking velocity ($Wv$) and stability ($St$), are the metrics for evaluating the performance of robot gait. Therefore, this problem is formulated as a two-objective optimization problem.

Walking velocity is an indicator that combines the speed and direction of the multi-legged robot. It is also suitable as an evolutionary fitness function. Speed can be calculated by capturing the Euclidean distance between the robot's start and end positions within a motion cycle, and the motion direction reflects the robot's ability to move straight in a specified direction. The specific definition is given as follows:
\begin{equation}\label{eq:wv}
Wv = 
\left\{
\begin{array}
{lr}
\zeta ^ s \cdot f_s + \zeta ^ q \cdot f_q , ~~~~~if~\Delta \mathbf{X}\cdot\mathbf{V}^*>0\\
-1, ~~~~~~~~~~~~~~~~~~~~~~~~~~~ otherwise\\
\end{array}
\right.
\end{equation}
$\zeta^s $ and $\zeta ^q$ are weight coefficients, $f_s$ indicates the speed in a motion cycle, $\Delta \mathbf{X}$ is the motion vector of the robot and $\mathbf{V}^*$ is a unit vector specifying the desired movement direction. We hope that the robot can move straight along the specified direction. Therefore, the walking efficiency is set to $-1$ if the robot moves in the opposite direction during walking, and all normal values of $Wv$ are considerably greater than $0$. The second term $f_q=exp(-\Delta \theta^2)$ reflects the deviation angle $\Delta \theta$ between the actual walking direction and the preset target direction. In the experiment we set the weight coefficients $\zeta^s $ and $\zeta ^q$ to 0.5.
Since the optimizer aims to minimize the objective function, the final objective of the velocity component is set as the reciprocal of $Wv$.

Stability is critical for robots to maintain balance and avoid hardware damage.
Accurate and efficient evaluation of stability is an important but non-trivial problem.
A standard solution is to directly use the stability score from the inertial measurement unit (IMU) as an indicator of the robot stability \cite{H2006Detection} \cite{Wang2009Modeling}.
However, researchers revealed that the use of the IMU readings only fails to accurately reflect the robot stability \cite{nygaard2017multi}. 
Here, we propose an objective function $St$ (Eq. \ref{eq:st3}), combining the information of linear acceleration and body orientation \cite{nygaard2017multi}.
Specifically, it is defined as:
\begin{align}
& F_j = \sqrt {\frac 1 N \sum _{i=1}^N (acc_{i,j}^2)- \overline{acc_j}^2} \label{eq:st1}\\
& G = \sqrt {\frac 1 N \sum _{i=1}^N roll_{i}^2}+\sqrt {\frac 1 N \sum _{i=1}^N pitch_{i}^2} \label{eq:st2}\\
& St = \frac {F_x+F_y+F_z} {scaling\_factor} + G \label{eq:st3}
\end{align}
where $acc, \overline{acc}$ are the current and average accelerations in a motion cycle, $i$ is the index of the sampling point (sampled at 100hz), and $j \in [x, y, z]$ is the axis. $roll$ and $pitch$ are the orientation values that are read from the sensor on the robot.  $roll$ is the orientation angle defined on the plane of $x$, $y$ axes, and $pitch$ is the orientation angle defined on the plane of $x$, $z$ axes.
The orientation component $G$ indicates that smaller values of $roll$ and $pitch$ reflect a preferred upright robot posture with higher stability. The integrative stability objective function shown in Eq.~\ref{eq:st3} is the sum of the linear acceleration and orientation components, as shown in Eq.~\ref{eq:st1} and Eq.~\ref{eq:st2}, respectively.
$scaling\_factor$ is used to balance the acceleration and direction so that they can evenly affect the fitness value, which is set to 50 in the experiment.



\subsection{Transfer Learning Modular}

We can use different transfer learning methods to achieve our goals. In this research, we use Transfer Component Analysis (TCA) \cite{pan2010domain} to conduct the experiments. The TCA method attempts to learn a set of potential common components that can be transferred among different domains. Therefore, the difference in data distribution between domains can be significantly reduced by mapping data from different domains to a latent space represented by the set of transferred components. Next, the standard machine learning classifier or regression model is trained across the domain in this latent space.
The TCA aims to find a better feature representation and reduce the difference of the data distribution between the source and target domains. Based on this consideration, the TCA uses a measure called maximum minimum distance (MMD) \cite{steinwart2001influence} to evaluate the differences between different domains, and it is defined as follows:
\begin{align}\label{eq:mmd}
  \mathbf{MMD} (\mathcal F,p,q)
  &:=\sum_{i=1}^m \sum_{j=1}^n \mathbf{tr}[\hat K (\frac 1 {m \times m} L_{ii}-\frac 1 {m \times n} L_{ij} \nonumber \\
  & - \frac 1 {n \times m} L_{ji} + \frac 1 {n \times n} L_{jj})] \\
  & := \mathbf{tr}(\hat K L) \nonumber
\end{align}
where $m$ and $n$ represent the number of samples in the source and target domains, $tr(A)$ represents the trace of matrix A, and the matrix $\hat{K}$ is defined as follows:
\begin{equation}\label{eq:K}
\hat K =
\left\{
  \begin{matrix}
   \hat K _{X,X} & \hat K _{X,Y} \\
   \hat K _{Y,X} & \hat K _{Y,Y}
  \end{matrix}
\right\} \in R^{(m+n)\times(m+n)}
\end{equation}
$\hat K _{X,Y}$ denotes the kernel matrix, its elements are $k_{i,j} = \kappa(x_i, y_i) = \phi(x_i)^T \phi(y_i)$, $\kappa(\cdot,\cdot)$ denotes the kernel function, $\phi(\cdot)$ denotes the feature mapping function. $\hat K _{X,Y}$ represents the data similarity in the domains $\mathbf X$ and $\mathbf Y$. The matrix $L$ contains the coefficients of the measurement matrix, the elements of which are as follows:
\begin{equation}\label{eq:L}
L(i,j) =
\left\{
  \begin{array}
  {lr}
  \frac 1 {m \times m}, x_i, x_j \in X \\
  \frac 1 {n \times n},  x_i, x_j \in Y \\
  - \frac 1 {m \times n}, \textrm{otherwise}
  \end{array}
\right.
\end{equation}

The optimization problem of the TCA algorithm can be written as:

\begin{equation}\label{eq:tcaopt}
  \begin{split}
  & \mathop{\arg} \mathop{\min}_W \ \mu \cdot \mathbf{tr}(W^TW)+\mathbf{tr} (W^TKLKW) \\
  & \textrm{subject to } W^TKHKW = \mathbf{I}
  \end{split}
\end{equation}

where in the case of $\mathbf H = \mathbf I - \frac 1 {m+n} \mathbf{11}^T$, $\mathbf I$ refers to the $(m + n) \times (m + n)$ identity matrix, $W^T W$ is a regularization term, and $\mathbf 1$ is a $(m + n) \times 1$ matrix whose elements are all $1$,  and $\mu$ is the trade-off coefficient.
The problem of solving the dimensionality reduction matrix W of the TCA algorithm can be converted into a matrix trace maximization problem. 
This problem can be solved by generalized eigenvalue decomposition (GED), in which the solution consists of the first $d$ eigenvectors sorted by the eigenvalue in a descending order \cite{mika1999fisher}. The TCA process is given in Algorithm 1.

\begin{algorithm}
  \caption{TCA}
  \label{alg:tca}
  \KwIn{Source domain $X$; target domain $Y$; kernel function $\kappa (\cdot, \cdot)$;}
  \KwOut{Matrix $W$}
  According to Equation~\ref{eq:K}-\ref{eq:tcaopt}, construct the Kernel Matrix $\hat K$, Matrix $L$, Matrix $H$;\\
  Construct Matrix $W$ by the $d$ leading eigenvectors of $(KLK+\mu I)^{-1}KHK$;\\
  \Return
      Matrix $W$;
\end{algorithm}

\begin{figure*}[htbp]
  \centering
  \includegraphics[width=18cm]{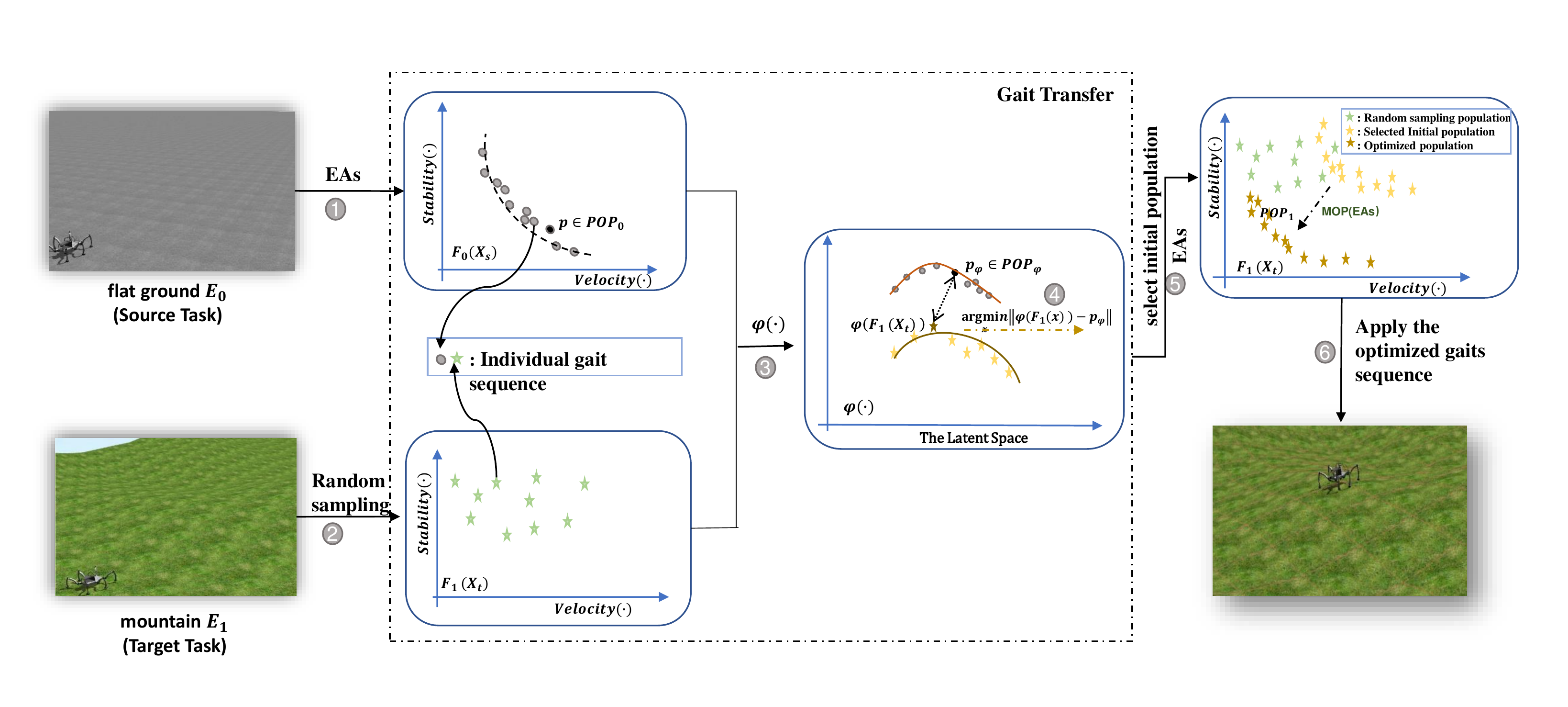}
  \caption{The pipeline of the proposed algorithm framework.}
  \label{fig:framework}
\end{figure*}

\subsection{Gait Transfer Algorithm Framework}

In this part, we propose an algorithm framework for robot gait optimization based on transfer learning, Tr-GO. 
We formulate the problem of robot gait generation as an MOP task and solve this by evolutionary computation. 
The technology of transfer learning uses existing experiences from the previous optimizations to improve the sampling quality of the initial population in new scenarios.
In this way, we hope that the algorithm framework can help us achieve better gait optimization results with less computational time and fewer resources. Meanwhile, the framework is universal and can be combined with most (if not any) population-based gait evolution algorithm.

\textbf{The motivation of this approach:} From the perspective of transfer learning, we believe that there is a certain correlation between gait optimization tasks in simple environments (such as flat ground) and those in more complex environments (such as rugged terrain or inclined slopes).
The gaits trained in simple and complex environments are regarded as the source and target domains, respectively. 
The robot gait  optimizations in simple and unknown complex environments are regarded as the source and target tasks, respectively.
These two tasks are similar in that they both attempt to generate a gait that maximally optimizes performance metrics given a specific environment. In the previous works, more attention was paid to the gait optimization task in a specific environment, and few works investigated how to utilize the correlation of the gait generation in different environments. 
This implies that by reusing the knowledge that had been learned in the past, the existing methods have significant room for performance boosting and reducing the consumption of computational resources.

The pipeline of the proposed algorithm framework is shown in Fig.~\ref{fig:framework}. Step 1 indicates that the robot gait is optimized by the EA in the source domain, and the optimized gait population set $POP_0$ in the source domain can be obtained. Our goal is to solve the task of gait generation and optimization in the new target environment (such as $E_1$), which requires random gait sampling in the new environment through Step 2. This step is closely related to the construction of the initial population of the EA.
The diversity and quality of the initial population determine the number of accessible features and eventually the success of gait optimization. 
Our work introduces transfer learning to use the set $POP_0$  in the source domain to assist the construction of the initial gait in the target domain. In the gait transfer module (Step 3), we choose transfer component analysis (TCA) \cite{pan2010domain} as the transfer learning tool of the algorithm framework. 
TCA maps different gait patterns in source and target domains to a hidden space with common transfer components, which can significantly reduce the difference of data distribution between domains. Through Step 4, we search for the population from the target domain in the hidden space to ensure that it is as close as possible to the population from the source domain. In Step 5, we generate an initial population for gait evolution in the target domain. This population retains the advantages of the optimized gait patterns from the source domain through transfer learning, and applies the knowledge gained from the source domain to the target task. 
Compared with the initial individuals generated by random sampling, the  ones generated by our method are expected to be closer to the Pareto front and shorten the evolution process. 
Furthermore, we will train the optimized population through the EA. Finally, in Step 6, we apply the optimized gait to the target environment. 
\begin{algorithm}[h]
	\caption{Tr-GO: Gait Optimization based on Transfer Learning}
	\label{alg:rtm}
	\KwIn{
		The optimized gait population of source domain $POP_S = \{x_1,...,x_m \}$, Gait population sampled from the target domain $X_T^E$ in  environment $E$; ~Optimization functions $F_S (\cdot)$ and $F_T (\cdot)$; ~ A kernel function $k(\cdot, \cdot)$ }
	\KwOut{Optimized gait sequences $POP_T$}
	Initialize\; 
	\While{a change occurred in target environment $E$}{
		$InitPOP_T =$ Tr-GIGP$(POP_S, X_T^E, k(\cdot, \cdot))$\;
		$InitPOP_T\xrightarrow[EAs]{MOP\left( F_T\left( x \right) \right)}POP_T$\;
		\Return
		$POP_T$\;	
	}
\end{algorithm}

The pseudo-code for the Tr-GO framework is shown in Algorithm 2. $POP_S$ is the optimized gait population in the source domain. $X_T^E$ is gait population sampled from the target domain in  environment $E$. Let $\left| POP_S \right|=n_s$ and $\left| X_T^E \right|=n_t$ represent the number of samples. $F_S(.), F_T(.)$ are the performance evaluation functions of the current robot gait. 
The Tr-GIGP algorithm outputs a population with no specialized constraints, so we can combine Tr-GIGP with most (if not all) population-based evolution algorithms (EAs). Similarly, we can replace TCA with other transfer methods in Tr-GIGP.
From the perspective of EC, the robot gait generation in a specific environment is equivalent to solving an MOP problem and finding the ultimate evolved population \cite{chang1999pareto}. When the task space is transfered to an unknown new environment, the optimization may be improved if it can reuse the optimized gait population (POP) in other similar scenarios.
When an environment change is detected, the algorithm triggers the transfer process.

\begin{algorithm}
	\caption{Tr-GIGP: Generate Initial Gait Population by using Transfer Learning }
	\label{alg:rtm}
	\KwIn{
		The optimized gait population of source domain $POP_S = \{x_1,...,x_m \}$, Gait population sampled from the target domain $X_T^E$ in  environment $E$; ~Optimization functions $F_S (\cdot)$ and $F_T (\cdot)$; ~ A kernel function $k(\cdot, \cdot)$ }
	\KwOut{Initial gait population after transfer $InitPOP_T$}
	Initialization\;
	Calculate the $F_S(POP_S)$ and $F_T(X_T^E)$\;
	$W\gets TCA\left(  F_S\left( POP_S \right) , F_T\left( X_T^E \right), \kappa \right)$\;
	$POP_\varphi \gets \emptyset$\;
	\For {$every ~ p \in POP_S$}{
		$ \kappa _p\gets \left[ \kappa \left( F_S\left( POP_S\left( 1 \right) \right) ,p \right) ,\cdots, \kappa \left( F_T\left( X_T^E\left( n_t \right) \right) ,p \right) \right]^{T} $\;	
		$\varphi \left( p \right) \gets W^T\kappa _p$\;
		$POP_\varphi \gets POP_\varphi \cup \left\{ \varphi \left( p \right) \right\}$\;
	}
	\For{$every ~p_\varphi \in POP_\varphi$}{
		$x\gets \underset{x}{arg\min}\lVert \varphi \left( F_T\left( x \right) \right) -p_{\varphi} \rVert$\;
		$InitPOP_T\gets InitPOP_T \cup \{x \}$\;
	}

	\Return
	$InitPOP_T$\;
\end{algorithm}


As the major part of the framework, Tr-GIGP (Algorithm 3) uses the optimized source gaits $POP_S$ and transfer learning to generate the initial gaits $InitPOP_T$ that can be applied to the target task. The initial phase (Line 1) efficiently evaluates the sampled population and removes some outliers, such as those that cannot enable the robot to maintain balance in the new environment. 
In Line 3 of Tr-GIGP algorithm, the data in the source and target domains are mapped to a common latent space using the TCA method, and in this space, the difference between the two distributions becomes as small as possible. Thus, we can use the $POP_S$ found in the simple environment $(E_0)$ (source domain) to generate the initial population in complex environments $(E_{1,2,...,n})$ for the evolutionary search of robot gait generation tasks. Note that the inputs to TCA are robot walking gait samples in a known simple environment $(E_0)$ and complex environments $(E_{1or2...n})$. The output of TCA is the transformation matrix $W$. We can use the matrix $W$ to construct the latent space. 
In Line 4 of Tr-GIGP, $POP_\varphi$ represents the set of particles in the latent space after mapping, which can be adjusted as a set of mapping solutions in the subsurface space. In Lines 10 to 13, we find $x$ such that $\varphi \left( F_T\left( x \right) \right)$ is the nearest neighbor to $p_\varphi \in POP_\varphi$. This indicates the demand to solve the single-objective optimization problem here, and we can apply any single-objective optimization algorithm to solve the problem. In this paper, we used the interior point algorithm provided by MATLAB to solve this problem. The calculated $x$ is the initial population $InitPOP_T$ to be used for gait evolution in the target environment. 

\section{Results \& Discussions}
\subsection{Experimental Settings}

The experiment in this work runs on the robot simulation platform, Webots \cite{michel2004cyberbotics}. 
The hexapod robot Mantis is provided by Micromagic Robot Laboratory.  
The detailed geometry and physical parameters of the robot are shown in Table I.
The initial values and the corresponding ranges of the joint angles are shown in Table II.

\begin{table}[ht]
	\centering
	\caption{Parameters of the mantis robot model}
	\begin{tabular}{p{6.945em}cc}
		\toprule
		Parameter & Body & Leg \\
		\midrule
		Mass(kg) & 5.2   & 0.9 \\
		\midrule
		Length(m) & 0.72  & 0.05 \\
		\midrule
		Width(m) & 0.42  & 0.05 \\
		\midrule
		Height(m) & 0.1   & 0.1 \\
		\bottomrule
	\end{tabular}%
	\label{tab:torso}%
\end{table}%
 
\begin{table}[h]
	\centering
	\caption{Parameters of the joint angles}
	\begin{tabular}{ccc}
		\toprule
		& range & initial \\
		\midrule
		\multicolumn{1}{c}{\multirow{2}[2]{*}{$\theta _1$}} & \multirow{2}[2]{*}{$-60^\circ\le \theta _1\le 20^\circ$}     & \multirow{2}[2]{*}{$-60^\circ$ (except Leg 3, 4 is $0^\circ$) } \\
		&       & \\
		\midrule
		\multicolumn{1}{c}{\multirow{2}[2]{*}{$\theta _2$}} & \multirow{2}[2]{*}{$-20^\circ\le \theta _2\le 40^\circ$} & \multirow{2}[2]{*}{$30^\circ$}\\
		&       &        \\
		\midrule
		\multicolumn{1}{c}{\multirow{2}[2]{*}{$\theta _3$}} & \multirow{2}[2]{*}{ $-140^\circ\le \theta _3\le -85^\circ$} & \multirow{2}[2]{*}{$-125^\circ$} \\
		&       &         \\      
		\bottomrule
	\end{tabular}%
	\label{tab:joint}%
\end{table}%

\begin{figure}[htbp]
  \centering
  \subfigure[source environment $E_0$]{\includegraphics[width=0.45\hsize]{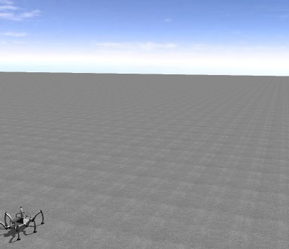}}
  \subfigure[target environment $E_1$]{\includegraphics[width=0.45\hsize]{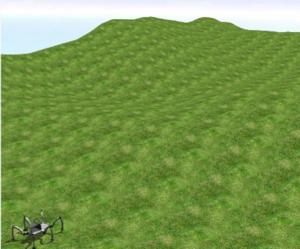}} \\
  \subfigure[target environment $E_2$]{\includegraphics[width=0.45\hsize]{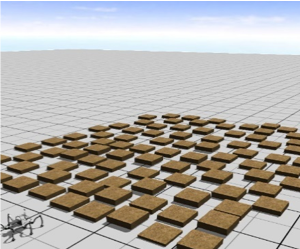}}
  \subfigure[target environment $E_3$]{\includegraphics[width=0.45\hsize]{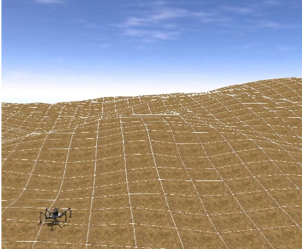}} \\
  \caption{Robots walk in different terrain environments.}
  \label{fig:environment}
\end{figure}

We simulate four types of terrain environments that robots may encounter. 
Gait optimization on the flat ground $E_0$ (Fig.~\ref{fig:environment}(a)) is regarded as the source domain.
We verify our method by transferring the knowledge gained from $E_0$ to three complex terrain environments, which were denoted as $E_1$, $E_2$, $E_3$. 
Fig.~\ref{fig:environment}(b) shows a grass terrain with inclined hills. This environment is complex in that the robot is prone to roll over and slip on the slope. It may be forced to turn when it fails to climb the hillside and keep straight walking.  Fig.~\ref{fig:environment}(c) shows that the obstacles are randomly placed on the ground to simulate rugged terrain, and the robot is prone to be tripped by an obstacle. Therefore, a gait is considered to be successful when it enables the robot to traverse the rugged terrain without being obstructed. The terrain in Fig.~\ref{fig:environment}(d) is similar to that in Fig.~\ref{fig:environment}(b), except for varying slope inclinations and shapes. 
For each target environment, the ability of navigating through the terrain to the target points specified in advance is regarded as success of robot motion generation. 
The surface inclination in the simulator is controlled by Berlin noise algorithm, where PerlinNoctaves (the number of octaves of the perlin noise) parameters corresponding to Fig.~\ref{fig:environment}(b) and Fig.~\ref{fig:environment}(d) are 3 and 5 respectively. 
The height of the rocks in  Fig.~\ref{fig:environment}(c) is set to between 0 and 0.5.

\begin{figure*}[bp]
	\centering
	\subfigure[]{\includegraphics[width=0.32\hsize]{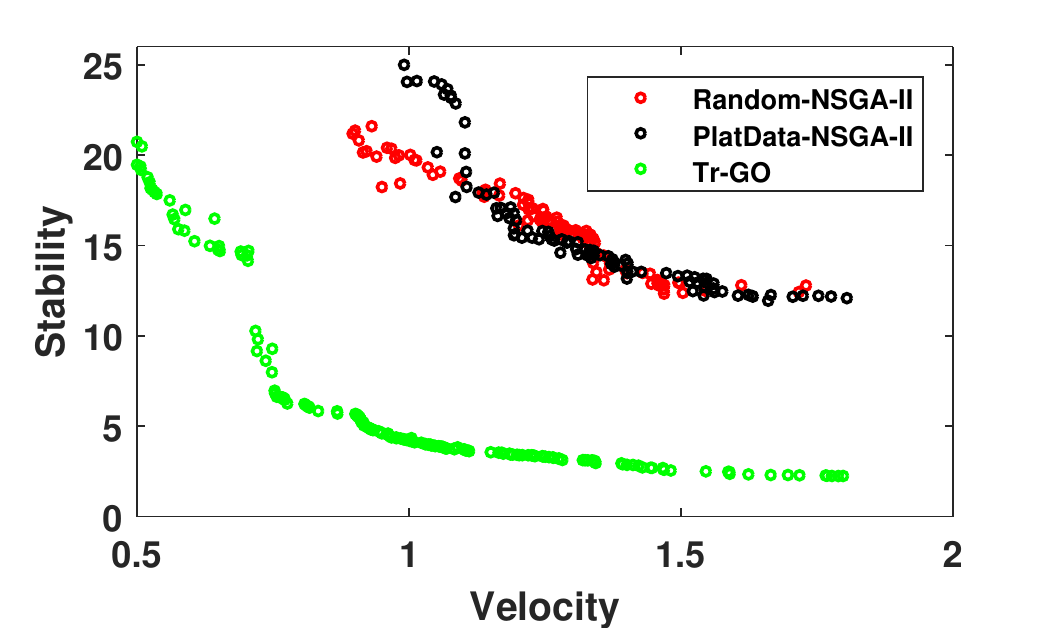}}
	\subfigure[]{\includegraphics[width=0.32\hsize]{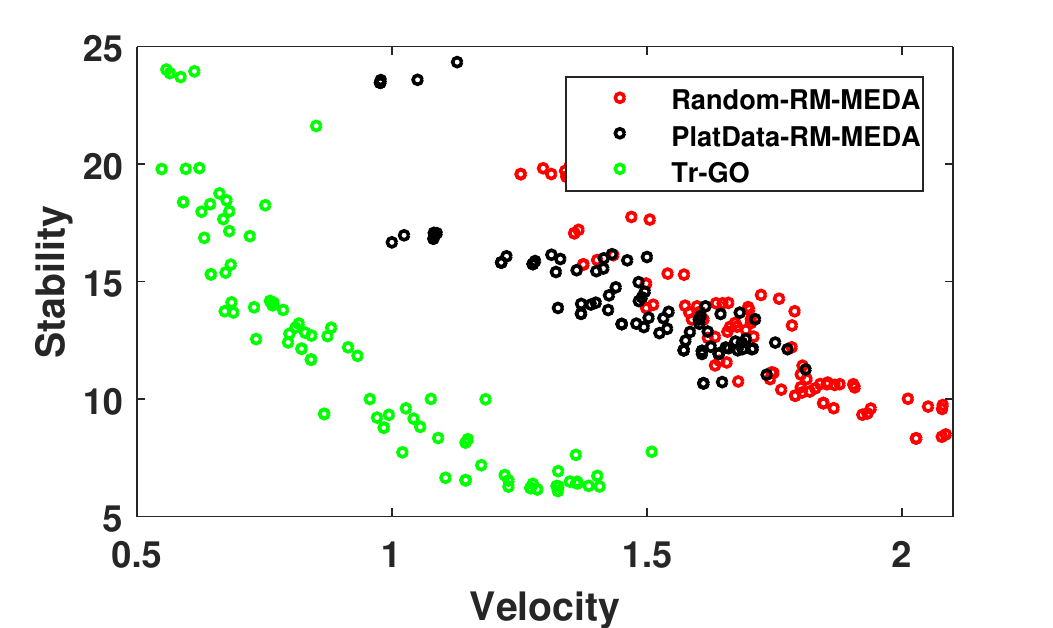}}
	\subfigure[]{\includegraphics[width=0.32\hsize]{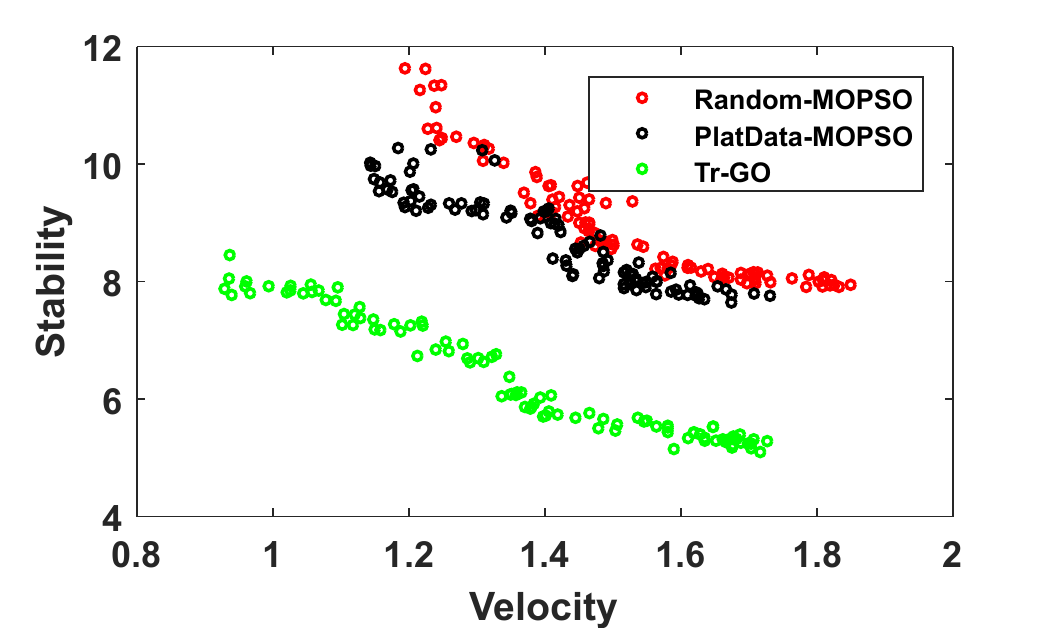}} \\
	\caption{Comparison of three EA algorithms in Environment 1: (a) NSGA-II, (b) RM-MEDA, and (c) MOPSO. Each algorithm is tested under three conditions: random sampling (colored in red), direct use of samples from $E_0$ (colored in black) and using our proposed Tr-Go framework (colored in green).}
	\label{fig:result_e1}
\end{figure*}

The Tr-GO framework proposed in this paper is compatible with any population-based optimization algorithm suitable for robot gait generation. As a case study, three representative multi-objective EAs - GA (NSGA-II) \cite{Kalyanmoy2002fast}, EDA (RM-MEDA) \cite{zhang2008rm}, PSO (MOPSO) \cite{coello2002mopso} - were selected in the experiment. 
The balance factor in the TCA algorithm is set to 0.5, the number of dimensions in the latent space is 20, and the Gaussian kernel function is used.

\subsection{Experimental Result}

We validate the effectiveness of our method in three target environments ($E_1, E_2, E_3$) and three different evolution algorithms (NSGA-II, RM-MEDA, MOPSO) in each environment.

\textbf{1) Experimental Results in Environment $E_1$}

First, the Tr-GO algorithm framework is tested using the NSGA-II algorithm. In Environment $E_0$, the NSGA-II algorithm iterates for 100 generations with an initial population of 200 randomly generated individuals.
After the optimization, the best 100 individuals are selected by the non-dominated sorting as the source domain. In Environment $E_1$, the motion generation task can be divided into three cases, each using a different multi-objective evolutionary algorithm. The first case (Random-NSGAII), as in Environment $E_0$, randomly samples 200 initial individuals in a complex terrain environment. In the second case (PlatData-NSGAII), the final gait optimization result $POP_S$ of the flat environment $E_0$ is \emph{directly used} as the initial population for the robot motion evolution in Environment $E_1$. In the third case (Tr-GO), the output after processing $POP_S$ with our Tr-GO algorithm framework serves as the initial population in Environment $E_1$. After that, 100 evolutionary iterations are performed for each of the three cases, and the top 100 individuals are selected according to the non-dominated sorting.

Fig.~\ref{fig:result_e1} (a) shows the final vector distribution of the objective function obtained under three scenarios. The individuals colored in red, black and green are generated by Random-NSGAII, PlatData-NSGAII and Tr-GO, respectively.
The intertwined distributions of red and black circles indicate that the robot gait performance obtained by Random-NSGAII and PlatData-NSGAII does not present significant difference after 100 iterations. 
Although the black and red circles represent different situations, the reason for the similar performance of the final optimization evaluation may be that the robot motion optimization from the flat terrain situation cannot be directly applied to the complex environment.
Without the process of domain transfer learning, \emph{direct use} of samples from the source domain is not different from randomly generating the initial population. 
The Pareto front of green circles is distinct from the distribution of the previous two, which means that the evaluation of the robot gait obtained by Tr-GO is significantly better than the other two. It confirms that the Tr-GO framework based on transfer learning can help NSGA-II to obtain the solution closer to the theoretical optimal boundary.

Similarly, we also test the effect of the Tr-GO framework using the RM-MEDA and MOPSO algorithms in the $E_1$ environment. The experiment is also divided into three cases for comparison. 
The final vector distribution of the objective function is shown in Fig.~\ref{fig:result_e1}(b) and Fig.~\ref{fig:result_e1}(c). 
The results show that the performance obtained by using the Tr-Go framework (colored in green) is better than the other two cases (colored in black and red, respectively).
\begin{figure*}[htbp]
	\centering
	\subfigure[Random-EA]{\includegraphics[width=0.32\hsize]{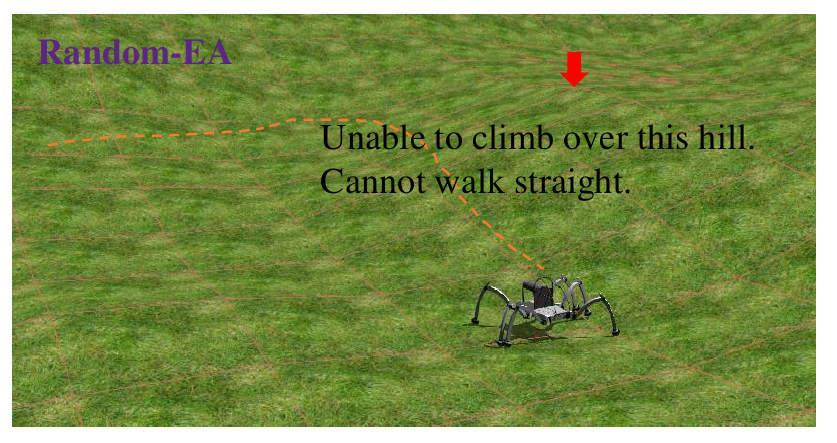}}
	\subfigure[PlatData-EA]{\includegraphics[width=0.32\hsize]{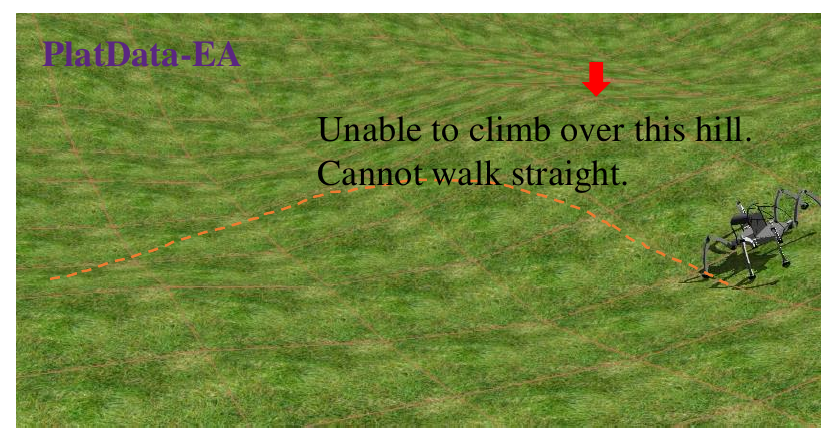}}
	\subfigure[Tr-Go]{\includegraphics[width=0.32\hsize]{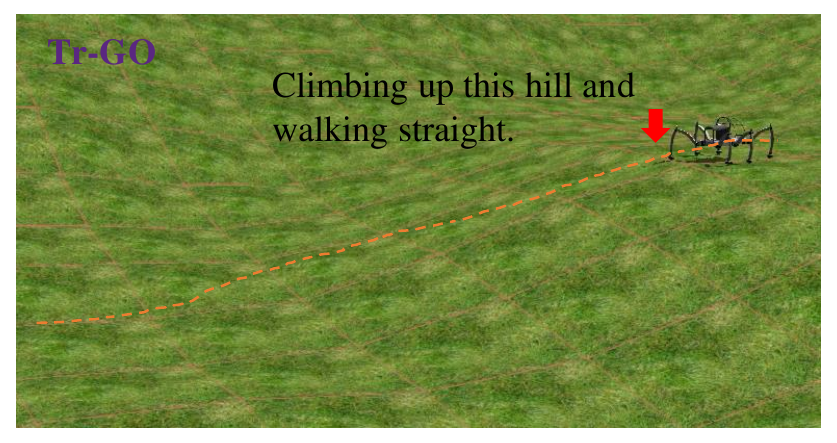}} \\
	\caption{Visual comparison of the robot walking in $E_1$ under three different scenarios. In (a) and (b), the robot is unable to climb over this hill and walk straight, while in (c), the robot successfully completes the task with our proposed Tr-Go framework.}
	\label{fig:visualcomp_e1}
\end{figure*}

In addition to the quantitative evaluation in Fig.~\ref{fig:result_e1}, we visually compare the walking performance of the robot in Environment $E_1$.
Fig.~\ref{fig:visualcomp_e1}(a) and (b) show that without the proposed Tr-GO algorithm framework, the robot fails to perform straight-line walking or climb over the hill when encountering the slopes.
Fig.~\ref{fig:visualcomp_e1}(c) demonstrates that the Tr-GO algorithm framework generates the optimal gait pattern and enables the robot to successfully climb over the hillside. 
This comparison confirms that the proposed Tr-Go framework improves the quality of individuals in the evolutionary computation.

\begin{figure*}[bp]
	\centering
	\subfigure[]{\includegraphics[width=0.32\hsize]{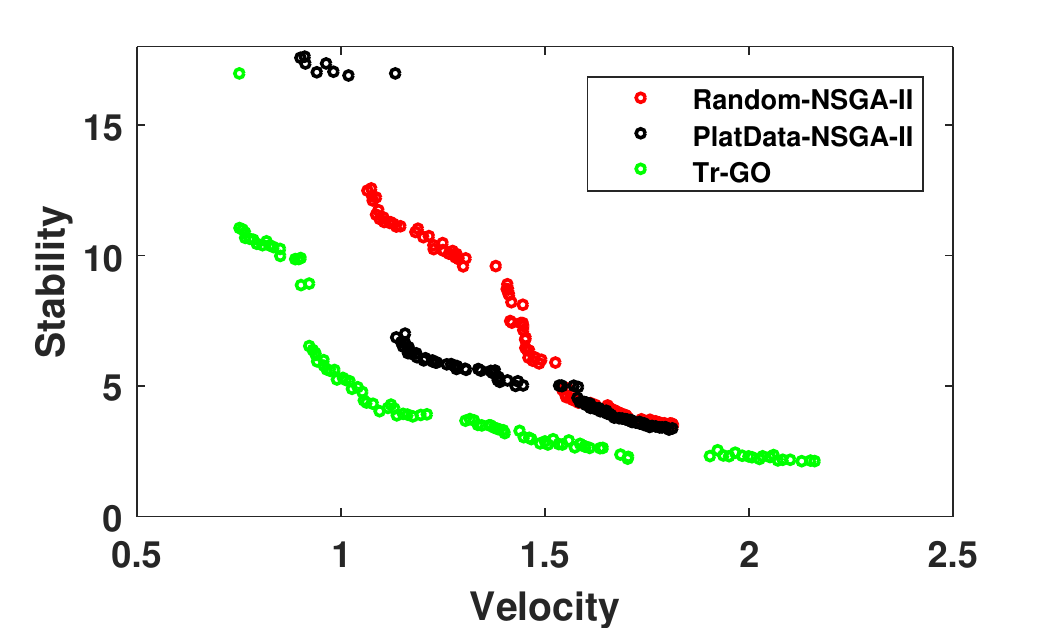}}
	\subfigure[]{\includegraphics[width=0.32\hsize]{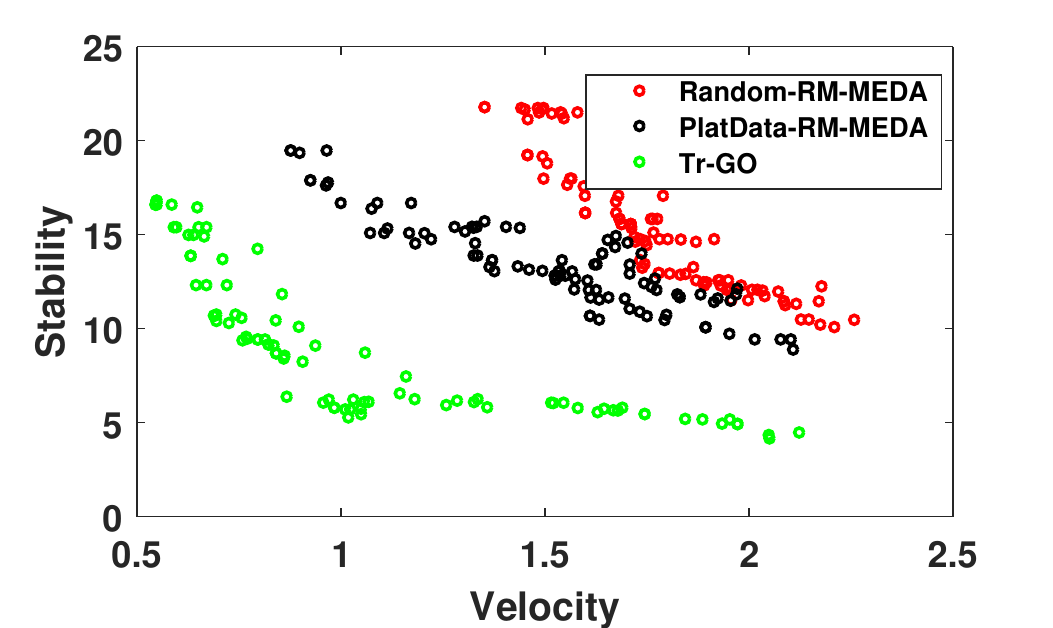}}
	\subfigure[]{\includegraphics[width=0.32\hsize]{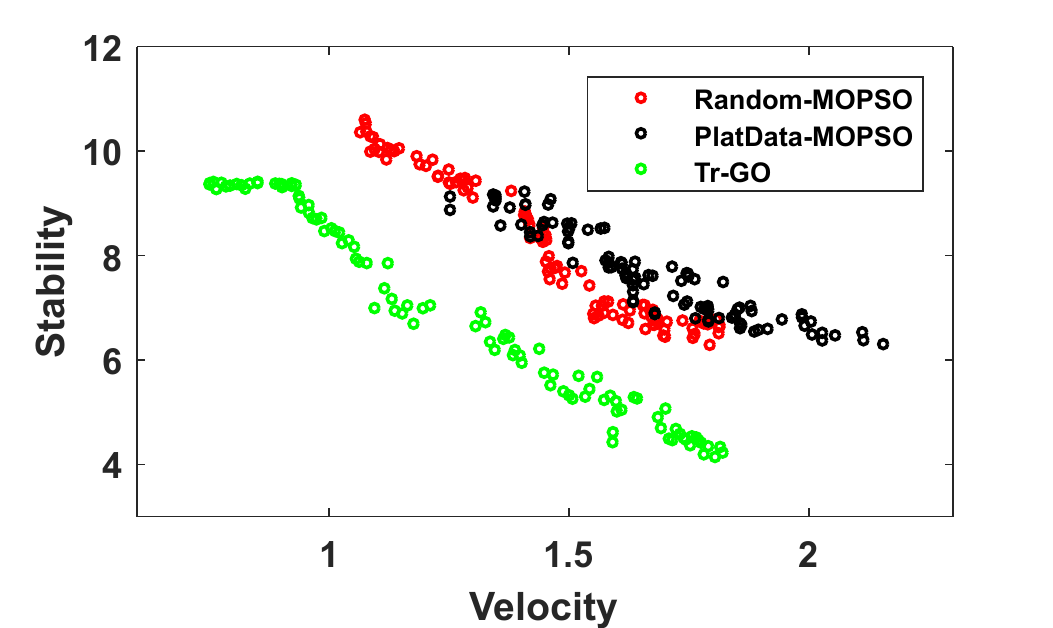}} \\
	\caption{Comparison of three EA algorithms in Environment 2: (a) NSGA-II, (b) RM-MEDA, and (c) MOPSO. Each algorithm is tested under three conditions: random sampling (colored in red), direct use of samples from $E_0$ (colored in black) and using our proposed Tr-Go framework (colored in green).}
	\label{fig:environment_2}
\end{figure*}


\textbf{2) Experimental Results in Environment $E_2$}

The experimental results in Environment $E_2$ (Fig.~\ref{fig:environment_2}) show that the performance of our Tr-GO framework (labeled as green circles) consistently outperforms that under the other two conditions (black and red circles).
This advantage is consistent across all three choices of EAs.
This robust performance confirms the merits of our method in producing high-quality initial population.

The difference from Environment $E1$ is that the direct use of the optimal solution in $E_0$ (PlatData-*) is slightly better than the standard strategy of random sampling (Random-*) (using NSGA-II Fig.~\ref{fig:environment_2}(a) and RM-MEDA Fig.~\ref{fig:environment_2}(b)). 
This indicates that in Environment $E_2$, using the optimization results from the flat terrain as the initial gait achieves a better performance than random gait generation.
The reason for this may be due to the fact that the $E_2$ environment has a direct correlation with the flat environment $E_0$ to some extent. 
An explanatory case is that the robot may repeatedly yet coincidentally step on an obstacle without falling into the gap between obstacles, creating an \emph{illusion} of walking on the flat terrain. 
This ideal example shows the direct similarities between walking on the flat terrain and traversing across the stacked obstacles.
Thus, the direct use of the optimization results of gait from the flat terrain can also improve the evolution result in the complex terrain $E_2$ to a minor extent.

When using the MOPSO algorithm (Fig.~\ref{fig:environment_2}(c)), the strategy of random gait generation even slightly outperforms that of the direct use of the optimal solution from $E_0$. We attribute this to the shortcomings of the MOPSO algorithm, such as its inability to maintain the diversity of solution sets and its tendency to fall into local optimal. The strategy of random gait generation may produce more diverse population, while the optimal solution from $E_0$ only represents a small sub-space of the complete sampling space.


\textbf{3) Experimental Results in Environment $E_3$}

\begin{figure*}[htbp]
	\centering
	\subfigure[]{\includegraphics[width=0.32\hsize]{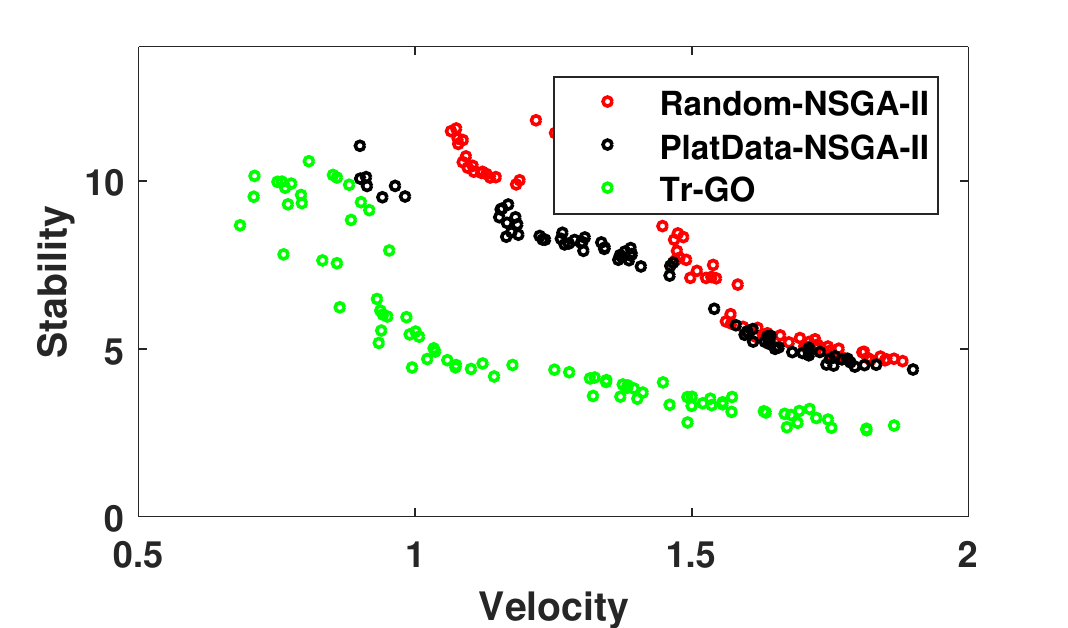}}
	\subfigure[]{\includegraphics[width=0.32\hsize]{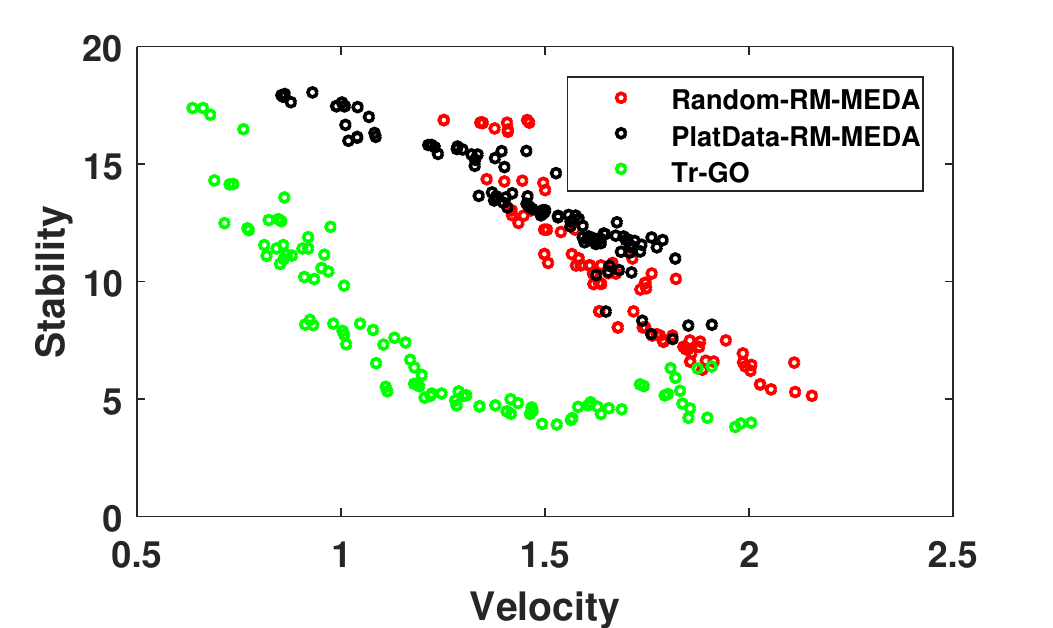}}
	\subfigure[]{\includegraphics[width=0.32\hsize]{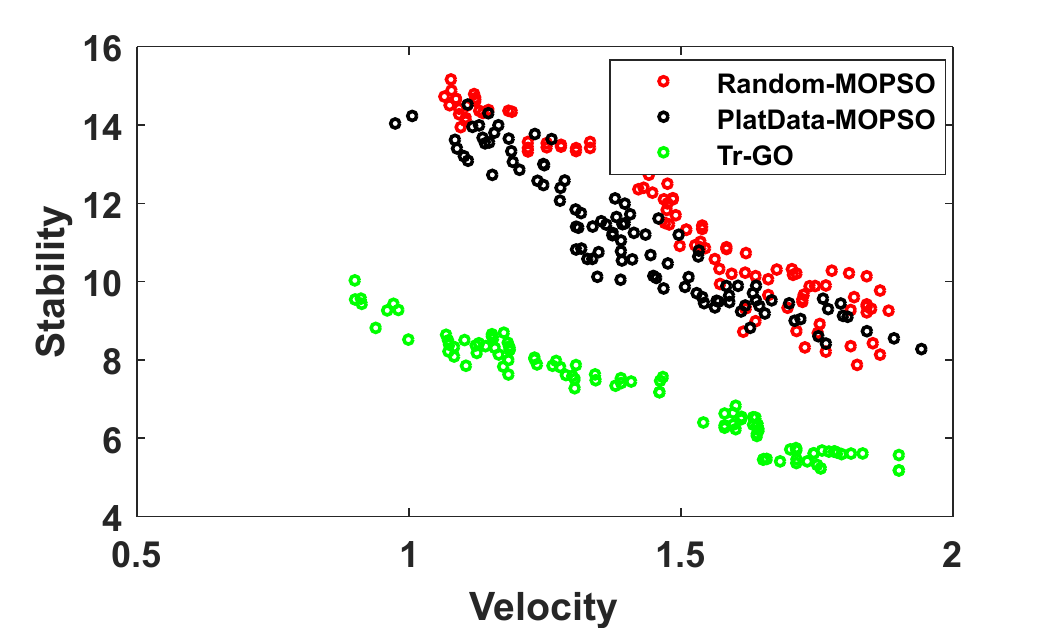}} \\
	\caption{Comparison of three EA algorithms in Environment 3: (a) NSGA-II, (b) RM-MEDA, and (c) MOPSO. Each algorithm is tested under three conditions: random sampling (colored in red), direct use of samples from $E_0$ (colored in black) and using our proposed Tr-Go framework (colored in green).}
	\label{fig:environment_3}
\end{figure*}
Fig.~\ref{fig:environment_3} presents the results from Environment $E_3$. Similarly, based on the three EA algorithms, the experimental results show that the Tr-GO framework can help to achieve better robot gaits, in terms of the metrics of stability and velocity. 
This advantage is consistent regardless of the exact choice of EA.

It is worth noting that the direct use of optimization results as the initial population from the flat terrain in complex environments is not always better than using a randomly generated population. For example, in Fig.~\ref{fig:environment_3}(b) in which the RM-MEDA algorithm is used, the evaluation result indicates that the black circles do not present a significant advantage over the red ones. 
One possible reason is that the difference between two environments, $E_0$ and $E_3$ does not offer the direct transfer between the two optimal populations. 
This is consistent with the results from Environment 1.
This reveals that the optimal posture used in slope terrains is different from that used in the flat ground, and presents larger deviations from that in the rugged ground (Environment 2).


\textbf{4) Evolutionary Efficiency}

\begin{figure*}[htbp]
	\centering
	\subfigure[]{\includegraphics[width=0.48\hsize]{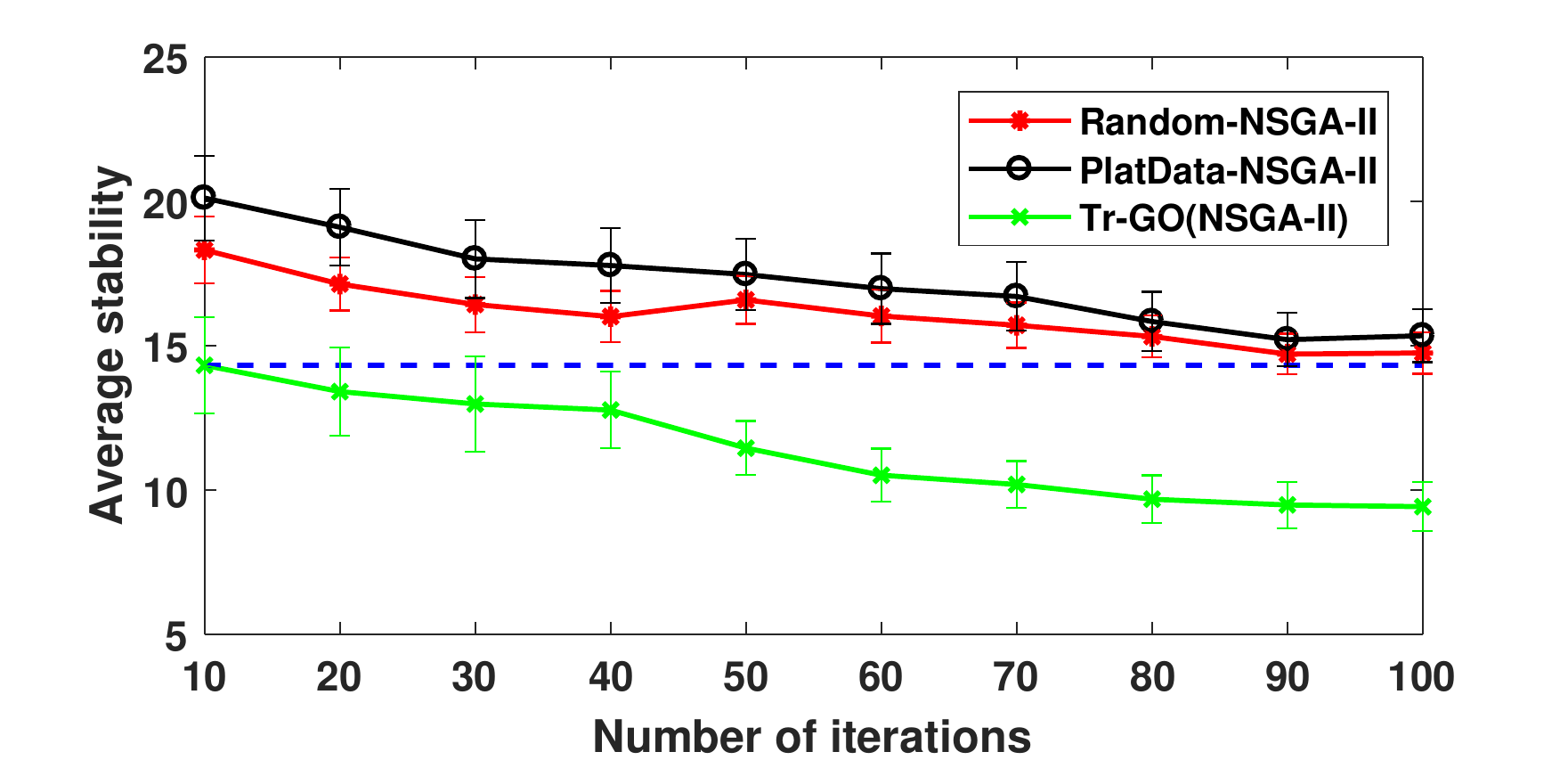}}
	\subfigure[]{\includegraphics[width=0.48\hsize]{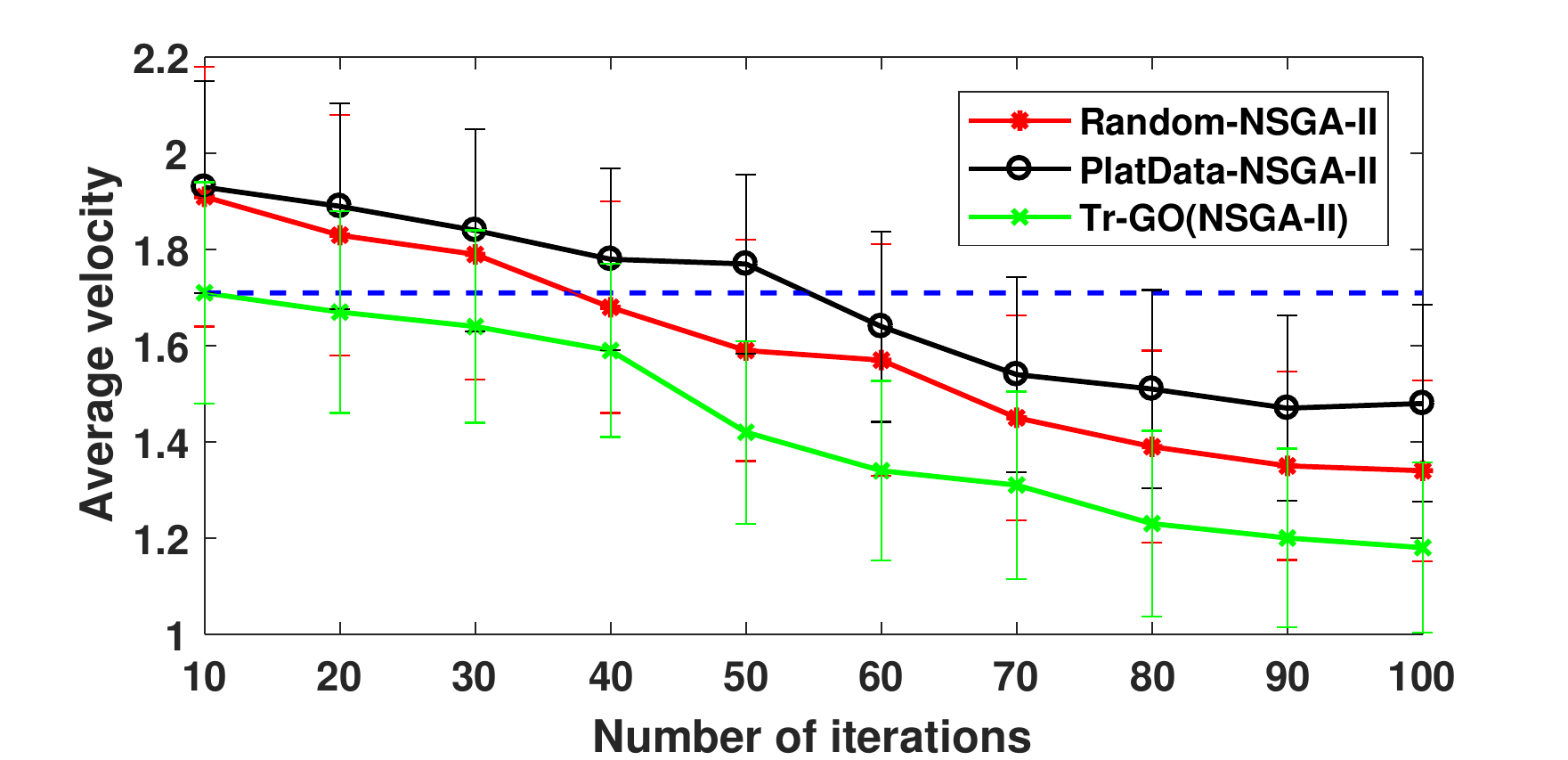}} \\
	\subfigure[]{\includegraphics[width=0.48\hsize]{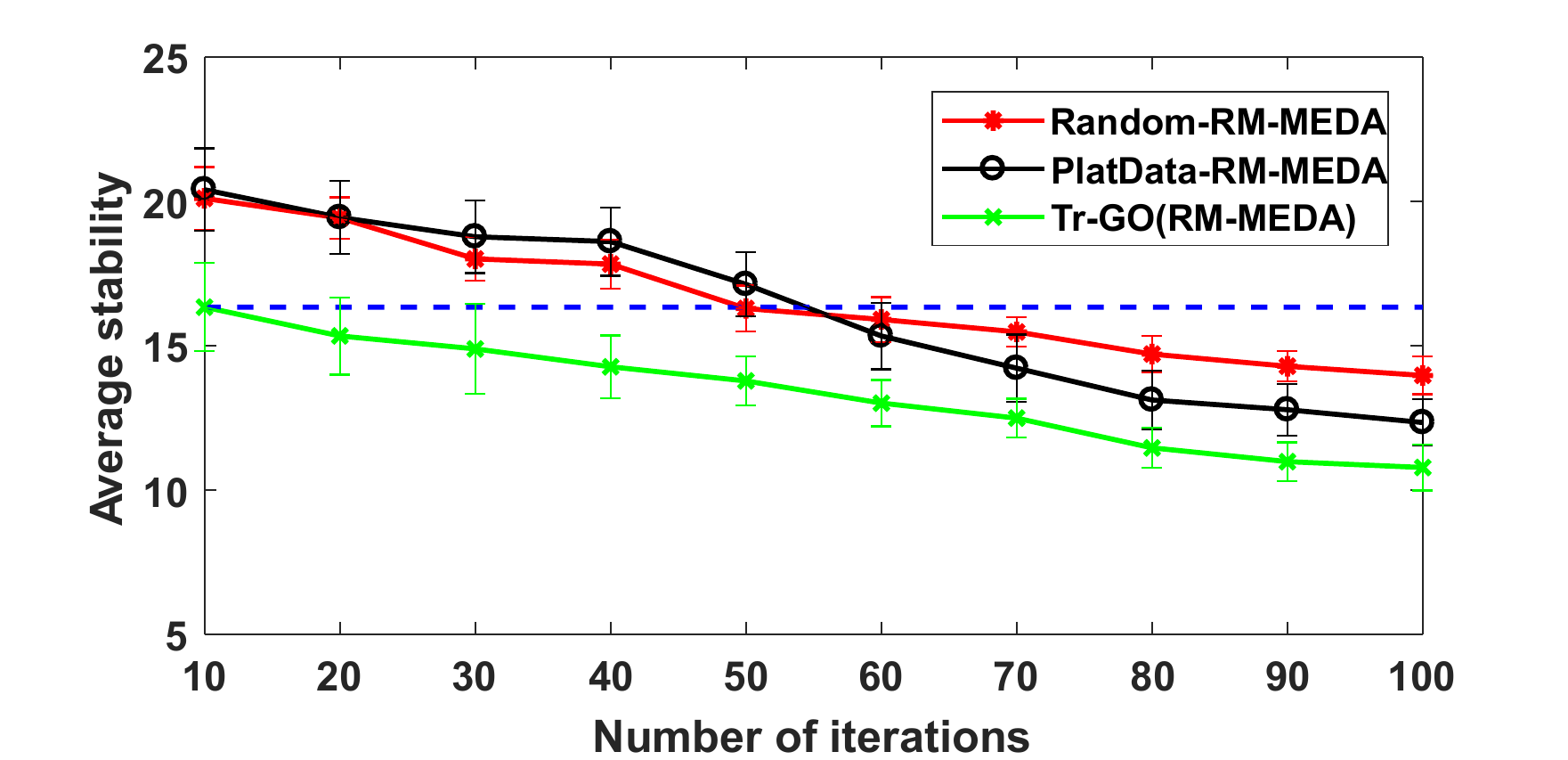}}
	\subfigure[]{\includegraphics[width=0.48\hsize]{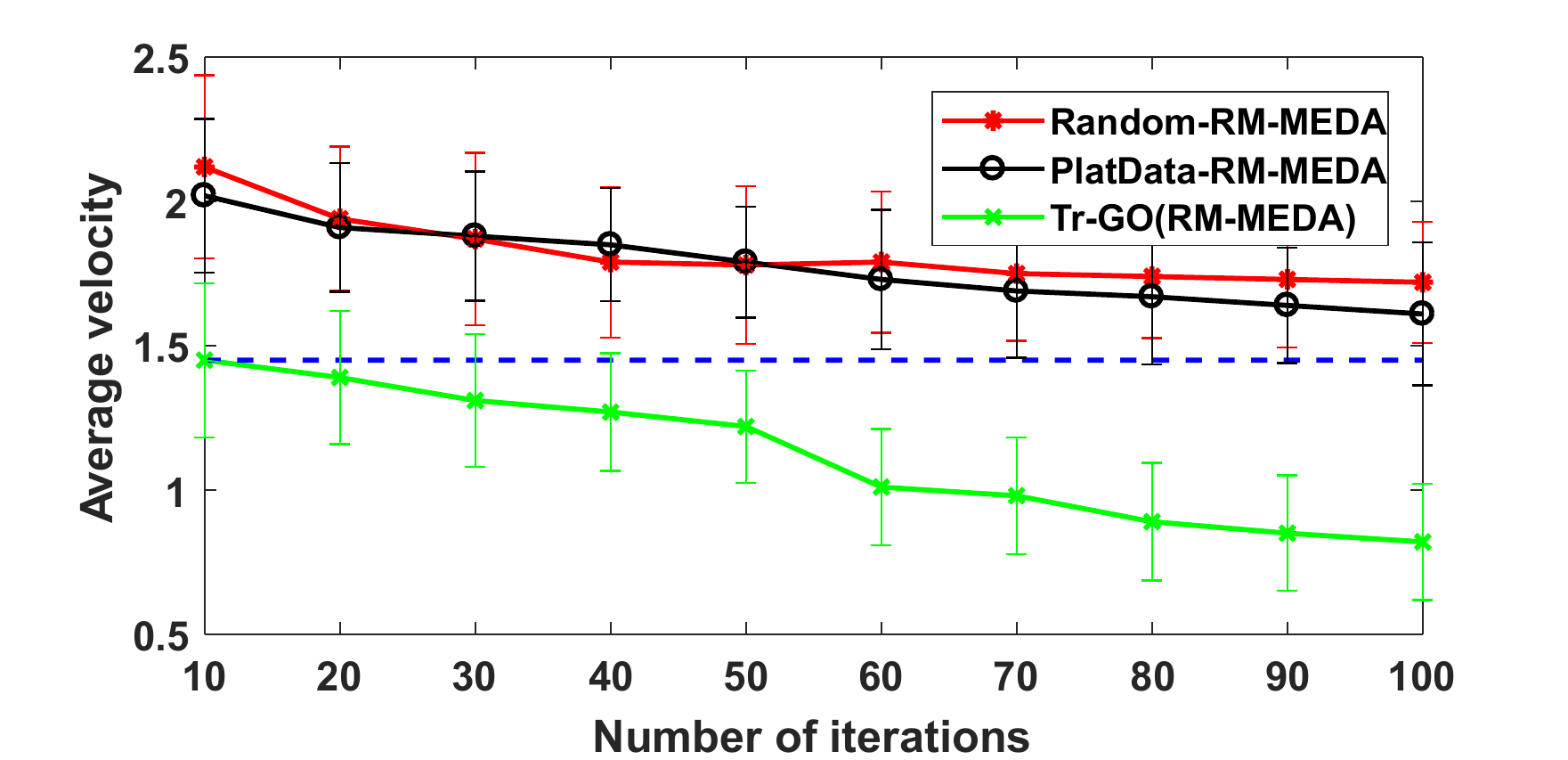}} \\
	\subfigure[]{\includegraphics[width=0.48\hsize]{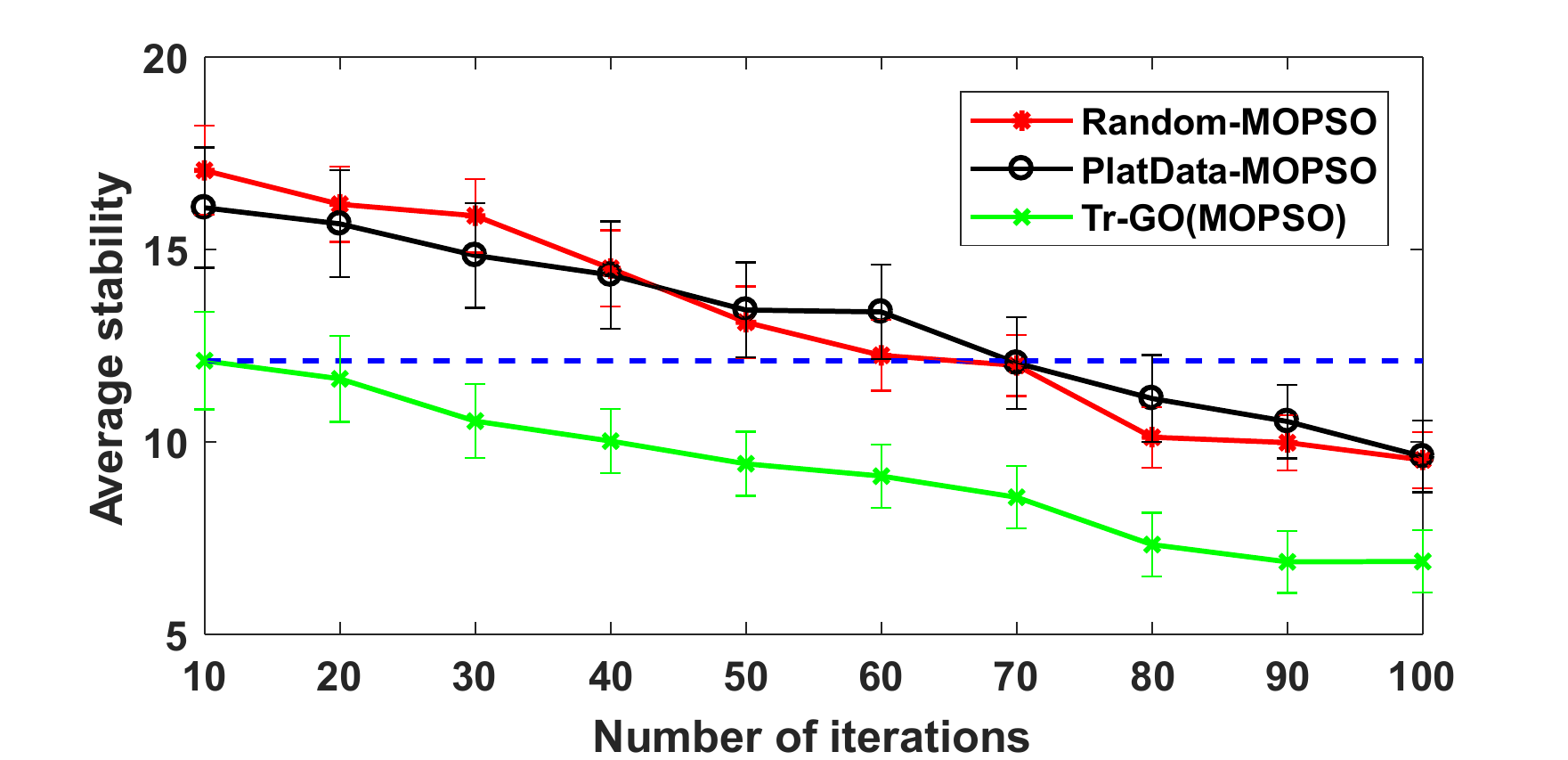}}
	\subfigure[]{\includegraphics[width=0.48\hsize]{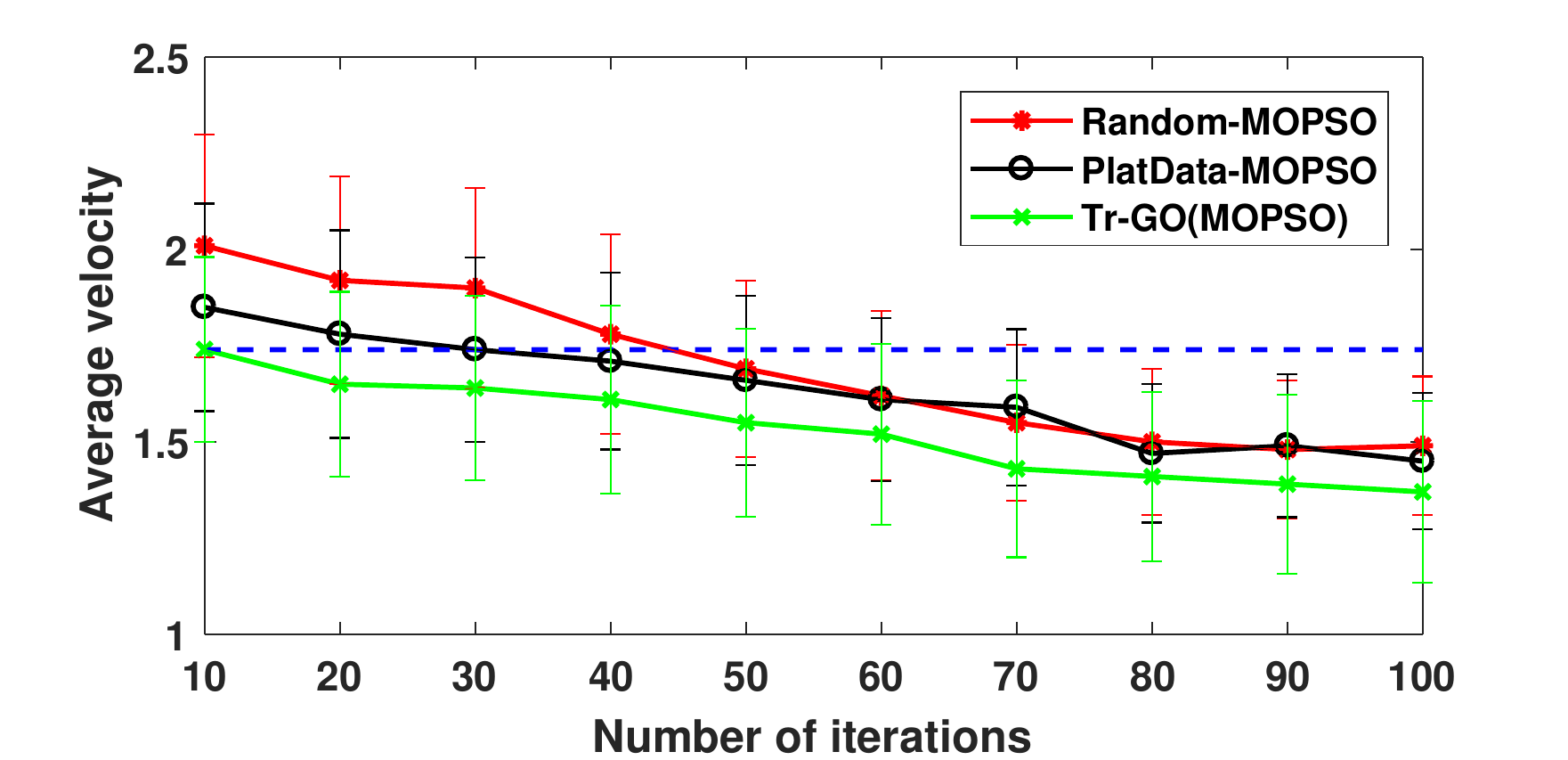}} \\
	\caption{The objective functions of stability (left column) and velocity (right column) during the evolution process. We compare the performance of these two metrics when using different algorithms. The blue line segment indicates the performance function value reached by the Tr-GO framework after the first ten generations of evolution.}
	\label{fig:evocurve}
\end{figure*}

We further investigate the influence of transfer learning on the evolution iteration in more details.
Fig.~\ref{fig:evocurve} plots the average values of the two objective functions in Environment $E_1$, under the three conditions mentioned above. 
The vertical axis is the average value of the  objective functions among the whole population. 
We use the value from our proposed Tr-Go framework after ten generations as the benchmark and compare the value with that of the other two conditions.
This benchmark is highlighted with a dashed horizontal line colored in blue.
The results show that Tr-GO possesses a performance advantage from the early stages of evolution and consistently maintains this advantage throughout the evolution process. 
We also compare the number of iterations to achieve the performance of the $10^{th}$ iteration using our Tr-GO framework. It can be seen that in (a)-(f), the other two conditions (Random-* and PlatData-*) require more evolutionary iterations to reach the same level of the performance metrics. 

We further compare the running time required to achieve the same performance (Table~\ref{tab:addlabel}). Specifically, we take the objective function values obtained after ten generations of the Tr-GO framework as the benchmark, and then record the time-cost for all three conditions to achieve this benchmark. 
The time required by Tr-GO evolution for ten generations is far less than that required by the other two conditions to achieve the same performance value. 
Our Tr-GO framework accelerates the evolution process by a minimum of 3-4 times compared with non-transferred scenarios.
For some cases, the baseline conditions never reach the benchmark, and we use ``$\infty$'' notation to indicate these failure cases.
In other words, our method initializes the evolution population with high quality and is capable of shortening the evolution process in an unknown new environment.

\begin{table*}[bp]
	\centering
	\begin{threeparttable}
		\caption{Time cost  (hours) for different approaches to achieve specified performance values}
		\begin{tabular}{|c|c|c|c|c|c|c|c|c|c|}
			\toprule
			\multicolumn{1}{|l|}{\multirow{2}[4]{*}{\textbf{}}} & \multicolumn{3}{c|}{\textbf{NSGA-II}} & \multicolumn{3}{c|}{\textbf{RM-MEDA}} & \multicolumn{3}{c|}{\textbf{MOPSO}} \\
			\cmidrule{2-10}          & Random & PlatDatas & \textbf{Tr-GO} & Random & PlatDatas & \textbf{Tr-GO} & Random & PlatDatas & \textbf{Tr-GO} \\
			\midrule
			\textbf{Stability} & 60.41 & $\infty$    & \textbf{7.04} & 31.22 & 34.29 & \textbf{6.24} & 53.28 & 22.86 & \textbf{7.61} \\
			\midrule
			\textbf{Velocity} & 28.42 & 42.26 & \textbf{7.68} & $\infty$    & $\infty$    & \textbf{6.81} & 21.97 & 32.98 & \textbf{7.33} \\
			\bottomrule
		\end{tabular}%
		\begin{tablenotes}
			\item[1] We take the objective functions values obtained after the Tr-GO algorithm framework has evolved for 10 generations as a standard.
			\item[2] The "$\infty$" in the table indicates that the specified performance index cannot be reached. 
		\end{tablenotes}
		
		\label{tab:addlabel}%
	\end{threeparttable}
\end{table*}%

\subsection{Limitation \& Failure Case}

While transfer learning in our framework improves the evolution performance of EC, they also present new challenges to our algorithms. When transfer learning techniques do not successfully construct the mapping between the source and target domains, they may even lead to negative transfer. 
The similarity between the source and target environments is the key factor in determining the effectiveness of the gait transfer.
We conducted further experiments by introducing a larger difference from Evironment $E_0$, such as setting PerlinNoctaves (defines the number of octaves of the perlin noise) parameters values in $E_1$ and $E_3$ to 6 to 11 (the higher the value, the more rugged the terrain is), or setting the height of rocks in $E_2$ to $\geqslant 0.6$.
The results show that our algorithm framework could not bring significant improvement in performance (Table~\ref{tab:addlabel4},~\ref{tab:addlabel5}). 
The reason may be that the difference between the target and source domains in these scenarios is too large to achieve effective gait transfer, which is a challenge to be solved in the future.

\begin{table*}[htbp]
	\centering
	\begin{threeparttable}
		\caption{Evolution performance of different perlin noise parameter values in a GA algorithm}
		\begin{tabular}{|c|c|c|c|c|c|c|c|c|c|}
			\toprule
			\multicolumn{1}{|l|}{\multirow{2}[4]{*}{\textbf{Perlin noise}}} & \multicolumn{3}{c|}{\textbf{6}} & \multicolumn{3}{c|}{\textbf{8}} & \multicolumn{3}{c|}{\textbf{11}} \\
			\cmidrule{2-10}          & Random & PlatDatas & \textbf{Tr-GO} & Random & PlatDatas & \textbf{Tr-GO} & Random & PlatDatas & \textbf{Tr-GO} \\
			\midrule
			\textbf{Stability} &15.57     & \textbf{15.21} & 16.32  & \textbf{19.98} & 20.34  & 26.33 & -1 & -1 & -1 \\
			\midrule
			\textbf{Velocity}  & 2.23 & \textbf{2.12} & 2.17 &  \textbf{2.38} & 2.45       & 3.63 & -1 & -1 & -1 \\		
			\bottomrule
			
		\end{tabular}%
		\begin{tablenotes}
			
			\item[1] The "\textbf{-1}" in the table indicates that the robot rolls over or reverses walking, and normal forward walking cannot be achieved.
		\end{tablenotes}
		
		\label{tab:addlabel4}%
	\end{threeparttable}
\end{table*}%

\begin{table*}[htbp]
	\centering
	\begin{threeparttable}
		\caption{Evolution performance of different height parameter values in a GA algorithm}
		\begin{tabular}{|c|c|c|c|c|c|c|c|c|c|}
			\toprule
			\multicolumn{1}{|l|}{\multirow{2}[4]{*}{\textbf{Hight}}} & \multicolumn{3}{c|}{\textbf{0.6}} & \multicolumn{3}{c|}{\textbf{0.8}} & \multicolumn{3}{c|}{\textbf{1.0}} \\
			\cmidrule{2-10}          & Random & PlatDatas & \textbf{Tr-GO} & Random & PlatDatas & \textbf{Tr-GO} & Random & PlatDatas & \textbf{Tr-GO} \\
			\midrule
			\textbf{Stability} & \textbf{29.34}     &31.98 & 32.33  & \textbf{43.33} & -1  &-1 & -1 & -1 & -1 \\
			\midrule
			\textbf{Velocity}  & \textbf{2.93} & 3.05 & 3.16 &  \textbf{4.45} & -1       & -1 & -1 & -1 & -1 \\		
			\bottomrule
		\end{tabular}%
		\begin{tablenotes}
			
			\item[1] The "\textbf{-1}" in the table indicates that the robot rolls over or reverses walking, and normal forward walking cannot be achieved.
		\end{tablenotes}
		
		\label{tab:addlabel5}%
	\end{threeparttable}
\end{table*}%

\section{Conclusion and Future Works}
Multi-legged robots have unique advantages in maintaining stability, but due to the high degrees of freedom, it is challenging to find a suitable gait pattern. 
This paper transforms the gait generation problem into a multi-objective optimization problem, and proposes an evolutionary multi-objective optimization framework based on transfer learning to solve this problem. The advantage of this solution is that the generated gait can not only adapt to different environments or tasks, but also meet multiple requirements simultaneously.
As such, these gaits can help a multi-legged robot better adapt to the environment and complete the specified tasks. 
This strategy of combining transfer learning and evolutionary algorithms can more effectively use the pre-acquired knowledge to generate an initial high-quality population.
Any kind of population-based optimization algorithms can be integrated into this framework without any further modification. 
This significantly improves the efficiency during the process of evolution computation and reduces the time cost to find the optimal Pareto front in the new environment.

There are a couple of directions for our future research. 
The proposed recipe of combining transfer learning and evolutionary computation not only can be used to solve the problem of gait generation, but also can be a promising solution in a larger scope.
Therefore, one of the future directions is to explore the application of this approach to the related problems in the field of robotics or even a wider range of domains.
We will further consider the integration of online transfer learning into this framework and enable the adaptation to achieve online transfer performance.
This could enable the robot to learn agile gait patterns when the environment is dynamically changing. 

\section*{Acknowledgment}
This work was supported in part by the National Natural Science Foundation of China under Grant 61673328 and Grant 61876162; in part by the Shenzhen Scientific Research and Development Funding Program under Grant JCYJ20180307123637294;

\bibliography{mybibtex}

\begin{thebibliography}{10}
\providecommand{\url}[1]{#1}
\csname url@samestyle\endcsname
\providecommand{\newblock}{\relax}
\providecommand{\bibinfo}[2]{#2}
\providecommand{\BIBentrySTDinterwordspacing}{\spaceskip=0pt\relax}
\providecommand{\BIBentryALTinterwordstretchfactor}{4}
\providecommand{\BIBentryALTinterwordspacing}{\spaceskip=\fontdimen2\font plus
\BIBentryALTinterwordstretchfactor\fontdimen3\font minus
  \fontdimen4\font\relax}
\providecommand{\BIBforeignlanguage}[2]{{%
\expandafter\ifx\csname l@#1\endcsname\relax
\typeout{** WARNING: IEEEtran.bst: No hyphenation pattern has been}%
\typeout{** loaded for the language `#1'. Using the pattern for}%
\typeout{** the default language instead.}%
\else
\language=\csname l@#1\endcsname
\fi
#2}}
\providecommand{\BIBdecl}{\relax}
\BIBdecl

\bibitem{todd2013walking}
D.~J. Todd, \emph{Walking machines: an introduction to legged robots}.\hskip
  1em plus 0.5em minus 0.4em\relax Springer Science \& Business Media, 2013.

\bibitem{jeong2019robust}
H.~Jeong, I.~Lee, J.~Oh, K.~K. Lee, and J.-H. Oh, ``A robust walking controller
  based on online optimization of ankle, hip, and stepping strategies,''
  \emph{IEEE Transactions on Robotics}, vol.~35, no.~6, pp. 1367--1386, 2019.

\bibitem{juang2017multiobjective}
C.-F. Juang and Y.-T. Yeh, ``Multiobjective evolution of biped robot gaits
  using advanced continuous ant-colony optimized recurrent neural networks,''
  \emph{IEEE transactions on cybernetics}, vol.~48, no.~6, pp. 1910--1922,
  2017.

\bibitem{boussema2019online}
C.~Boussema, M.~J. Powell, G.~Bledt, A.~J. Ijspeert, P.~M. Wensing, and S.~Kim,
  ``Online gait transitions and disturbance recovery for legged robots via the
  feasible impulse set,'' \emph{IEEE Robotics and Automation Letters}, vol.~4,
  no.~2, pp. 1611--1618, 2019.

\bibitem{gong2010review}
D.~Gong, J.~Yan, and G.~Zuo, ``A review of gait optimization based on
  evolutionary computation,'' \emph{Applied Computational Intelligence and Soft
  Computing}, vol. 2010, 2010.

\bibitem{chernova2004evolutionary}
S.~Chernova and M.~Veloso, ``An evolutionary approach to gait learning for
  four-legged robots,'' in \emph{2004 IEEE/RSJ International Conference on
  Intelligent Robots and Systems (IROS)(IEEE Cat. No. 04CH37566)},
  vol.~3.\hskip 1em plus 0.5em minus 0.4em\relax IEEE, 2004, pp. 2562--2567.

\bibitem{calandra2016bayesian}
R.~Calandra, A.~Seyfarth, J.~Peters, and M.~P. Deisenroth, ``Bayesian
  optimization for learning gaits under uncertainty,'' \emph{Annals of
  Mathematics and Artificial Intelligence}, vol.~76, no. 1-2, pp. 5--23, 2016.

\bibitem{rai2018bayesian}
A.~Rai, R.~Antonova, S.~Song, W.~Martin, H.~Geyer, and C.~Atkeson, ``Bayesian
  optimization using domain knowledge on the atrias biped,'' in \emph{2018 IEEE
  International Conference on Robotics and Automation (ICRA)}.\hskip 1em plus
  0.5em minus 0.4em\relax IEEE, 2018, pp. 1771--1778.

\bibitem{degrave2015transfer}
J.~Degrave, M.~Burm, P.-J. Kindermans, J.~Dambre, and F.~Wyffels, ``Transfer
  learning of gaits on a quadrupedal robot,'' \emph{Adaptive Behavior},
  vol.~23, no.~2, pp. 69--82, 2015.

\bibitem{wright2015intelligent}
J.~Wright and I.~Jordanov, ``Intelligent approaches in locomotion-a review,''
  \emph{Journal of Intelligent \& Robotic Systems}, vol.~80, no.~2, pp.
  255--277, 2015.

\bibitem{he2019survey}
J.~He, J.~Shao, G.~Sun, and X.~Shao, ``Survey of quadruped robots coping
  strategies in complex situations,'' \emph{Electronics}, vol.~8, no.~12, p.
  1414, 2019.

\bibitem{8100935}
M.~{Jiang}, Z.~{Huang}, L.~{Qiu}, W.~{Huang}, and G.~G. {Yen}, ``Transfer
  learning-based dynamic multiobjective optimization algorithms,'' \emph{IEEE
  Transactions on Evolutionary Computation}, vol.~22, no.~4, pp. 501--514,
  2018.

\bibitem{weiss2016survey}
K.~Weiss, T.~M. Khoshgoftaar, and D.~Wang, ``A survey of transfer learning,''
  \emph{Journal of Big data}, vol.~3, no.~1, p.~9, 2016.

\bibitem{9097186}
M.~{Jiang}, Z.~{Wang}, L.~{Qiu}, S.~{Guo}, X.~{Gao}, and K.~C. {Tan}, ``A fast
  dynamic evolutionary multiobjective algorithm via manifold transfer
  learning,'' \emph{IEEE Transactions on Cybernetics}, pp. 1--12, 2020.

\bibitem{marler2004survey}
R.~T. Marler and J.~S. Arora, ``Survey of multi-objective optimization methods
  for engineering,'' \emph{Structural and multidisciplinary optimization},
  vol.~26, no.~6, pp. 369--395, 2004.

\bibitem{pareto1906manuale}
V.~Pareto, ``Manuale di economia politica, milano,'' \emph{Societ{\`a} editrice
  libraria}, 1906.

\bibitem{Athan1996A}
T.~W. Athan and P.~Y. Papalambros, ``A note on weighted criteria methods for
  compromise solutions in multi-objective optimization,'' \emph{Engineering
  Optimization}, vol.~27, no.~2, pp. 155--176, 1996.

\bibitem{deb2001multi}
K.~Deb, \emph{Multi-objective optimization using evolutionary
  algorithms}.\hskip 1em plus 0.5em minus 0.4em\relax John Wiley \& Sons, 2001,
  vol.~16.

\bibitem{mcneill2002energetics}
R.~McNeill~Alexander, ``Energetics and optimization of human walking and
  running: the 2000 raymond pearl memorial lecture,'' \emph{American journal of
  human biology}, vol.~14, no.~5, pp. 641--648, 2002.

\bibitem{bertram2005constrained}
J.~E. Bertram, ``Constrained optimization in human walking: cost minimization
  and gait plasticity,'' \emph{Journal of experimental biology}, vol. 208,
  no.~6, pp. 979--991, 2005.

\bibitem{Manoj2006Computer}
S.~Manoj and R.~Andy, ``Computer optimization of a minimal biped model
  discovers walking and running,'' \emph{Nature}, vol. 439, no. 7072, pp.
  72--75, 2006.

\bibitem{hereid2018dynamic}
A.~Hereid, C.~M. Hubicki, E.~A. Cousineau, and A.~D. Ames, ``Dynamic humanoid
  locomotion: A scalable formulation for hzd gait optimization,'' \emph{IEEE
  Transactions on Robotics}, vol.~34, no.~2, pp. 370--387, 2018.

\bibitem{silva2016open}
F.~Silva, M.~Duarte, L.~Correia, S.~M. Oliveira, and A.~L. Christensen, ``Open
  issues in evolutionary robotics,'' \emph{Evolutionary computation}, vol.~24,
  no.~2, pp. 205--236, 2016.

\bibitem{takemori2018gait}
T.~Takemori, M.~Tanaka, and F.~Matsuno, ``Gait design for a snake robot by
  connecting curve segments and experimental demonstration,'' \emph{IEEE
  Transactions on Robotics}, vol.~34, no.~5, pp. 1384--1391, 2018.

\bibitem{nolfi2016evolutionary}
S.~Nolfi, J.~Bongard, P.~Husbands, and D.~Floreano, ``Evolutionary robotics,''
  in \emph{Springer Handbook of Robotics}.\hskip 1em plus 0.5em minus
  0.4em\relax Springer, 2016, pp. 2035--2068.

\bibitem{floreano2008evolutionary}
D.~Floreano, P.~Husbands, and S.~Nolfi, ``Evolutionary robotics,''
  \emph{Springer handbook of robotics}, pp. 1423--1451, 2008.

\bibitem{doncieux2013behavioral}
S.~Doncieux and J.-B. Mouret, ``Behavioral diversity with multiple behavioral
  distances,'' in \emph{2013 IEEE Congress on Evolutionary Computation}.\hskip
  1em plus 0.5em minus 0.4em\relax IEEE, 2013, pp. 1427--1434.

\bibitem{tellez2006evolving}
R.~A. T{\'e}llez, C.~Angulo, and D.~E. Pardo, ``Evolving the walking behaviour
  of a 12 dof quadruped using a distributed neural architecture,'' in
  \emph{International Workshop on Biologically Inspired Approaches to Advanced
  Information Technology}.\hskip 1em plus 0.5em minus 0.4em\relax Springer,
  2006, pp. 5--19.

\bibitem{bongard2008accelerating}
J.~C. Bongard, ``Accelerating self-modeling in cooperative robot teams,''
  \emph{IEEE Transactions on Evolutionary Computation}, vol.~13, no.~2, pp.
  321--332, 2008.

\bibitem{Cardenasmaciel2011Generation}
S.~L. Cardenasmaciel, O.~Castillo, and L.~T. Aguilar, ``Generation of walking
  periodic motions for a biped robot via genetic algorithms,'' vol.~11, no.~8,
  pp. 5306--5314, 2011.

\bibitem{eaton2015evolutionary}
M.~Eaton, \emph{Evolutionary Humanoid Robotics}.\hskip 1em plus 0.5em minus
  0.4em\relax Springer, 2015.

\bibitem{gupta2018trajectory}
G.~Gupta and A.~Dutta, ``Trajectory generation and step planning of a 12 dof
  biped robot on uneven surface,'' \emph{Robotica}, vol.~36, no.~7, pp.
  945--970, 2018.

\bibitem{sarkar20158}
A.~Sarkar and A.~Dutta, ``8-dof biped robot with compliant-links,''
  \emph{Robotics and Autonomous Systems}, vol.~63, pp. 57--67, 2015.

\bibitem{cardenas2011generation}
S.~L. Cardenas-Maciel, O.~Castillo, and L.~T. Aguilar, ``Generation of walking
  periodic motions for a biped robot via genetic algorithms,'' \emph{Applied
  Soft Computing}, vol.~11, no.~8, pp. 5306--5314, 2011.

\bibitem{lim2014gait}
I.-s. Lim, O.~Kwon, and J.~H. Park, ``Gait optimization of biped robots based
  on human motion analysis,'' \emph{Robotics and autonomous Systems}, vol.~62,
  no.~2, pp. 229--240, 2014.

\bibitem{wang2017autonomous}
W.~Wang, D.~Gu, and G.~Xie, ``Autonomous optimization of swimming gait in a
  fish robot with multiple onboard sensors,'' \emph{IEEE Transactions on
  Systems, Man, and Cybernetics: Systems}, vol.~49, no.~5, pp. 891--903, 2017.

\bibitem{jiang2017motion}
M.~Jiang, Z.~Huang, G.~Jiang, M.~Shi, and X.~Zeng, ``Motion generation of
  multi-legged robot in complex terrains by using estimation of distribution
  algorithm,'' in \emph{2017 IEEE Symposium Series on Computational
  Intelligence (SSCI)}.\hskip 1em plus 0.5em minus 0.4em\relax IEEE, 2017, pp.
  1--6.

\bibitem{9262190}
Z.~{Zhao}, G.~e.~{Zhang}, M.~{Jiang}, L.~{Feng}, and K.~C. {Tan}, ``Ednas: An
  efficient neural architecture design based on distribution estimation,'' in
  \emph{2020 2nd International Conference on Industrial Artificial Intelligence
  (IAI)}, 2020, pp. 1--6.

\bibitem{nygaard2017multi}
T.~F. Nygaard, J.~Torresen, and K.~Glette, ``Multi-objective evolution of fast
  and stable gaits on a physical quadruped robotic platform,'' in \emph{2017
  IEEE Symposium Series on Computational Intelligence (SSCI)}, 2017, pp. 1--8.

\bibitem{kobayashi2015selection}
T.~Kobayashi, T.~Aoyama, K.~Sekiyama, and T.~Fukuda, ``Selection algorithm for
  locomotion based on the evaluation of falling risk,'' \emph{IEEE Transactions
  on Robotics}, vol.~31, no.~3, pp. 750--765, 2015.

\bibitem{raj2019multiobjective}
M.~Raj, V.~B. Semwal, and G.~Nandi, ``Multiobjective optimized bipedal
  locomotion,'' \emph{International Journal of Machine Learning and
  Cybernetics}, vol.~10, no.~8, pp. 1997--2013, 2019.

\bibitem{Nasu2002Optimal}
Y.~Nasu, ``Optimal trajectory generation for a prismatic joint biped robot
  using genetic algorithms,'' \emph{Robotics Autonomous Systems}, vol.~38,
  no.~2, pp. 119--128, 2002.

\bibitem{moore2016comparison}
J.~M. Moore and P.~K. McKinley, ``A comparison of multiobjective algorithms in
  evolving quadrupedal gaits,'' in \emph{International Conference on Simulation
  of Adaptive Behavior}.\hskip 1em plus 0.5em minus 0.4em\relax Springer, 2016,
  pp. 157--169.

\bibitem{Kalyanmoy2002fast}
K.~Deb, A.~Pratap, S.~Agarwal, and T.~Meyarivan, ``A fast and elitist
  multiobjective genetic algorithm: Nsga-ii, ieee trans. on evol,'' \emph{IEEE
  Transactions on Evolutionary Computation - TEC}, vol.~6, 01 2002.

\bibitem{pugh2016quality}
J.~K. Pugh, L.~B. Soros, and K.~O. Stanley, ``Quality diversity: A new frontier
  for evolutionary computation,'' \emph{Frontiers in Robotics and AI}, vol.~3,
  p.~40, 2016.

\bibitem{cully2013behavioral}
A.~Cully and J.-B. Mouret, ``Behavioral repertoire learning in robotics,'' in
  \emph{Proceedings of the 15th annual conference on Genetic and evolutionary
  computation}.\hskip 1em plus 0.5em minus 0.4em\relax ACM, 2013, pp. 175--182.

\bibitem{cully2016evolving}
------, ``Evolving a behavioral repertoire for a walking robot,''
  \emph{Evolutionary computation}, vol.~24, no.~1, pp. 59--88, 2016.

\bibitem{cully2015robots}
A.~Cully, J.~Clune, D.~Tarapore, and J.-B. Mouret, ``Robots that can adapt like
  animals,'' \emph{Nature}, vol. 521, no. 7553, p. 503, 2015.

\bibitem{Duarte2018Evolution}
M.~Duarte, J.~Gomes, S.~M. Oliveira, and A.~L. Christensen, ``Evolution of
  repertoire-based control for robots with complex locomotor systems,''
  \emph{IEEE Transactions on Evolutionary Computation}, vol.~22, no.~2, pp.
  314--328, 2018.

\bibitem{9122031}
M.~{JIANG}, Z.~{WANG}, H.~{HONG}, and G.~G. {YEN}, ``Knee point based
  imbalanced transfer learning for dynamic multi-objective optimization,''
  \emph{IEEE Transactions on Evolutionary Computation}, pp. 1--1, 2020.

\bibitem{9185798}
Z.~{Zhao}, M.~{Jiang}, S.~{Guo}, Z.~{Wang}, F.~{Chao}, and K.~C. {Tan},
  ``Improving deep learning based optical character recognition via neural
  architecture search,'' in \emph{2020 IEEE Congress on Evolutionary
  Computation (CEC)}, 2020, pp. 1--7.

\bibitem{9199822}
M.~{Jiang}, Z.~{Wang}, S.~{Guo}, X.~{Gao}, and K.~C. {Tan}, ``Individual-based
  transfer learning for dynamic multiobjective optimization,'' \emph{IEEE
  Transactions on Cybernetics}, pp. 1--14, 2020.

\bibitem{9002942}
Z.~{WANG}, M.~{JIANG}, X.~{GAO}, L.~{FENG}, W.~{HU}, and K.~C. {TAN},
  ``Evolutionary dynamic multi-objective optimization via regression transfer
  learning,'' in \emph{2019 IEEE Symposium Series on Computational Intelligence
  (SSCI)}, 2019, pp. 2375--2381.

\bibitem{michel2004cyberbotics}
O.~Michel, ``Cyberbotics ltd. webots™: professional mobile robot
  simulation,'' \emph{International Journal of Advanced Robotic Systems},
  vol.~1, no.~1, p.~5, 2004.

\bibitem{iwasa2016motion}
M.~Iwasa, T.~Obo, and N.~Kubota, ``Motion generation of multi-legged robot by
  using knowledge transfer in rough terrain,'' in \emph{2016 IEEE Symposium
  Series on Computational Intelligence (SSCI)}.\hskip 1em plus 0.5em minus
  0.4em\relax IEEE, 2016, pp. 1--5.

\bibitem{guo2014locomotion}
S.~Guo, J.~Chang, X.~Yang, W.~Wang, and J.~Zhang, ``Locomotion skills for
  insects with sample-based controller,'' in \emph{Computer Graphics Forum},
  vol.~33, no.~7.\hskip 1em plus 0.5em minus 0.4em\relax Wiley Online Library,
  2014, pp. 31--40.

\bibitem{H2006Detection}
O.~H\"{o}hn, J.~Ga\v{c}nik, and W.~Gerth, ``Detection and classification of
  posture instabilities of bipedal robots,'' in \emph{Climbing \& Walking
  Robots- International Conference on Climbing \& Walking Robots \& the Support
  Technologies for Mobile Machines}, 2006.

\bibitem{Wang2009Modeling}
H.~Wang, J.~Zhang, J.~Yi, D.~Song, S.~Jayasuriya, and J.~Liu, ``Modeling and
  motion stability analysis of skid-steered mobile robots,'' in \emph{IEEE
  International Conference on Robotics \& Automation}, 2009.

\bibitem{pan2010domain}
S.~J. Pan, I.~W. Tsang, J.~T. Kwok, and Q.~Yang, ``Domain adaptation via
  transfer component analysis,'' \emph{IEEE Transactions on Neural Networks},
  vol.~22, no.~2, pp. 199--210, 2010.

\bibitem{steinwart2001influence}
I.~Steinwart, ``On the influence of the kernel on the consistency of support
  vector machines,'' \emph{Journal of machine learning research}, vol.~2, no.
  Nov, pp. 67--93, 2001.

\bibitem{mika1999fisher}
S.~Mika, G.~Ratsch, J.~Weston, B.~Scholkopf, and K.-R. Mullers, ``Fisher
  discriminant analysis with kernels,'' in \emph{Neural networks for signal
  processing IX: Proceedings of the 1999 IEEE signal processing society
  workshop (cat. no. 98th8468)}.\hskip 1em plus 0.5em minus 0.4em\relax Ieee,
  1999, pp. 41--48.

\bibitem{chang1999pareto}
C.-S. Chang, D.~Xu, and H.~Quek, ``Pareto-optimal set based multiobjective
  tuning of fuzzy automatic train operation for mass transit system,''
  \emph{IEE Proceedings-Electric Power Applications}, vol. 146, no.~5, pp.
  577--583, 1999.

\bibitem{zhang2008rm}
Q.~Zhang, A.~Zhou, and Y.~Jin, ``Rm-meda: A regularity model-based
  multiobjective estimation of distribution algorithm,'' \emph{IEEE
  Transactions on Evolutionary Computation}, vol.~12, no.~1, pp. 41--63, 2008.

\bibitem{coello2002mopso}
C.~C. Coello and M.~S. Lechuga, ``Mopso: A proposal for multiple objective
  particle swarm optimization,'' in \emph{Proceedings of the 2002 Congress on
  Evolutionary Computation. CEC'02 (Cat. No. 02TH8600)}, vol.~2.\hskip 1em plus
  0.5em minus 0.4em\relax IEEE, 2002, pp. 1051--1056.

\end{thebibliography}

\end{document}